\pdfoutput=1
\documentclass{article}

\usepackage{microtype}
\usepackage{graphicx}
\usepackage{subcaption}
\usepackage{booktabs}
\usepackage{algorithm}
\usepackage{algorithmic}
\usepackage{hyperref}
\usepackage{xurl}

\newif\ifshowauthors
\showauthorstrue
\ifshowauthors
  \usepackage[preprint]{icml2026}
\else
  \usepackage{icml2026}
\fi

\usepackage{amsmath}
\usepackage{amssymb}
\usepackage{mathtools}
\usepackage{amsthm}
\usepackage{multirow}
\usepackage[table]{xcolor}
\usepackage[capitalize,noabbrev]{cleveref}

\theoremstyle{plain}
\newtheorem{theorem}{Theorem}[section]
\newtheorem{proposition}[theorem]{Proposition}

\theoremstyle{definition}

\theoremstyle{remark}

\newcommand{\cmark}{\checkmark}
\definecolor{oursbg}{RGB}{232,240,252}
\definecolor{modelbg}{gray}{0.93}
\definecolor{avgbg}{gray}{0.955}
\newcommand{\dm}[1]{\textcolor{black!42}{#1}}   %

\usepackage{xspace}

\usepackage{placeins}
\newsavebox{\apbox}
\newlength{\apwd}
\newcommand{\fittab}[2]{%
  \sbox{\apbox}{#2}%
  \setlength{\apwd}{\wd\apbox}%
  \ifdim\apwd<0.55\textwidth\setlength{\apwd}{0.55\textwidth}\fi
  \begin{minipage}{\apwd}%
  \caption{#1}%
  \centering\usebox{\apbox}%
  \end{minipage}}


\icmltitlerunning{DForkTree: Accelerating Speculative Decoding with Fork-Aware Block Diffusion Draft Trees}

\begin{document}
\hypersetup{pdfsubject={Preprint}}

\twocolumn[
  \icmltitle{DForkTree: Accelerating Speculative Decoding with\\ Fork-Aware Block Diffusion Draft Trees}

  \icmlsetsymbol{equal}{*}
  \icmlsetsymbol{corr}{$\dagger$}
  \ifshowauthors
    \begin{icmlauthorlist}
      \icmlauthor{Zhicheng Zheng}{equal,hkustgz}
      \icmlauthor{Cheng Zhang}{equal,hkustgz}
      \icmlauthor{Chenkai Tang}{thu}
      \icmlauthor{Tianxiang Zhao}{corr,hkustgz}
    \end{icmlauthorlist}
    \vskip 0.04in
    {\centering\normalsize
     \textsuperscript{1}The Hong Kong University of Science and Technology (Guangzhou)\\
     \textsuperscript{2}Tsinghua University\\
     zzheng186@connect.hkust-gz.edu.cn\quad tianxiangz@hkust-gz.edu.cn\par}
    \vskip 0.03in
    {\centering\normalsize\textbf{Code:} \href{https://github.com/Perseus037/DForkTree}{https://github.com/Perseus037/DForkTree}\par}
  \else
    \begin{icmlauthorlist}
      \icmlauthor{Anonymous Authors}{anon}
    \end{icmlauthorlist}
    \icmlaffiliation{anon}{Anonymous Institution}
    \icmlcorrespondingauthor{Anonymous Authors}{anon@example.com}
  \fi
  \icmlkeywords{Speculative Decoding, Draft Trees, Large Language Models, Efficient Inference}
  \vskip 0.17in
]

\ifshowauthors
  \makeatletter\global\icml@noticeprintedtrue\makeatother
  {\let\thefootnote\relax\footnotetext{\hspace*{-\footnotesep}\textsuperscript{*}Equal contribution. \textsuperscript{$\dagger$}Corresponding author.}}
\else
  \printAffiliationsAndNotice{}
\fi

\begin{abstract}
Speculative decoding speeds up language models by having a lightweight drafter propose future tokens that the target model verifies in parallel.
Block diffusion drafters such as DFlash predict an entire block in one forward pass, and DDTree builds draft trees directly from their per-position distributions.
Each distribution implicitly assumes that the drafter's own greedy continuation is accepted, whereas a draft tree improves over a chain only on branches that leave this greedy path.
Once a path forks, the drafter's top-1 accuracy drops from 89\% to 18\% and its top candidates often miss the true token.
Re-running the drafter from the fork token largely restores its accuracy, showing that the errors stem from conditioning on an outdated prefix rather than from limited model capacity.
In this paper, we introduce DForkTree, a training-free method that re-anchors each branch on its own tokens.
It adjusts the drafter's scores with pointwise mutual information from target-generated $n$-grams, weighted more heavily after a fork, and a layer-wise GPU search recovers the exact optimal tree under these scores.
On Qwen3.5-4B and 9B, DForkTree outperforms all baselines on every benchmark, by up to 41.5\% over DFlash and 15.1\% over DDTree, with average speedups of up to $4.75\times$ over autoregressive decoding.
\end{abstract}

\section{Introduction}
\label{sec:intro}

\begin{figure}[t]
  \centering
  \includegraphics[width=\linewidth]{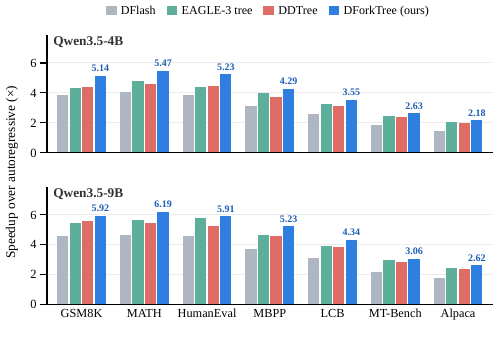}
  \vspace{-3mm}
  \caption{Speedup over autoregressive decoding at temperature 0 on seven benchmarks (MATH is MATH-500). DFlash drafts a 24-token chain, and all trees use its drafter.}
  \label{fig:speedup}
\end{figure}

\begin{figure*}[t]
  \centering
  \includegraphics[width=\textwidth]{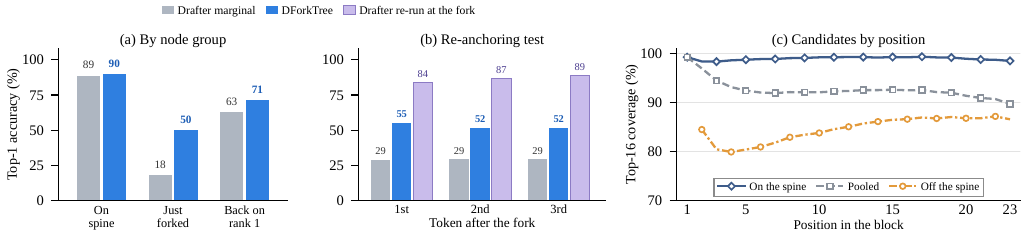}
  \vspace{-3mm}
  \caption{Where the marginals fail on Qwen3.5-4B. (a) Top-1 accuracy of the drafter's marginal and of DForkTree's score by node group. (b) Top-1 accuracy at the first three tokens after the first fork, with the drafter re-run from the fork token. (c) Share of positions whose true token is in the top 16 while all earlier ones are.}
  \label{fig:motivation}
  \vspace{2mm}
\end{figure*}

Large language models (LLMs) generate text one token at a time, and each step spends most of its time reading the model weights from memory.
Speculative decoding uses the spare computation by letting a lightweight drafter propose several future tokens that the target model verifies in one forward pass \citep{leviathan2023fast,chen2023accelerating}.
Because the target accepts only the tokens that it would have produced itself, the output distribution does not change.
Draft trees verify many branching continuations at once \citep{miao2024specinfer,cai2024medusa,li2024eagle2,chen2024sequoia}, and the extra nodes help only when some of these branches are accepted.

Block diffusion drafters such as DFlash \citep{dflash2026} make drafting much faster by predicting all positions of a block in a single forward pass.
In exchange, each position receives a marginal distribution that is not conditioned on the tokens a branch has chosen at earlier positions.
DDTree \citep{ringel2026ddtree} builds draft trees directly from the product of these marginals, and later methods add learned modules to restore the missing dependencies between positions \citep{ubtree2026,graft2026,dartree2026,oda2026weaver,dominotree2026}.
A common premise across these works is that the drafter already holds the right tokens among its top candidates and only fails to rank them \citep{dflash2blog2026,eisenstadt2026dcoco}.
None of these works, however, measures where in a tree the marginals go wrong, leaving it unclear which nodes need correction.

We call the drafter's greedy continuation its \emph{spine} and follow the branches that fork away from it.
Because each marginal averages over the prefixes that the drafter considers likely, it is shaped mostly by the spine.
This is a poor fit for tree drafting, because a tree gains over a chain only on branches that leave the spine.
On the spine, the marginals give a correct top-1 token in 89\% of cases.
Right after a path forks, top-1 accuracy falls to 18\%, up to a quarter of the true tokens drop out of the top 16, and the path probabilities overestimate acceptance.
To test whether this drop reflects limited capacity, we re-anchor the drafter by re-running it from the fork token as the next round would.
Its top-1 accuracy on the following token then rises from 29\% inside the original block to 84\%, so the drafter predicts the branch well once it is conditioned on the branch's prefix.
The same split accounts for up to 95\% of the coverage decay that prior work attributes to the drafter's capacity \citep{dflash2blog2026}.

Re-anchoring needs the target's features of the branch, which arrive only in the next round, so we introduce DForkTree, a training-free method that approximates it within the current round.
Using $n$-grams counted offline in target-generated text, it conditions each candidate on the tokens that precede it along its own branch.
Bayes' rule shows that this context should enter the drafter's score as pointwise mutual information (PMI).
Its weight follows how much the $n$-grams add beyond the drafter and doubles after a fork.
A layer-wise GPU search finds the exact optimal tree under these scores in one CUDA graph.

We evaluate DForkTree on Qwen3.5-4B and 9B \citep{qwen35} on seven benchmarks.
It is the fastest tree method on every benchmark (\cref{fig:speedup}), with up to 9.9\% higher throughput than the strongest tree baseline, 15.1\% higher than DDTree and 41.5\% higher than the DFlash chain, and it decodes up to $4.75\times$ faster than autoregressive decoding on average.
Our main contributions are as follows.
\vspace{-7pt}
\begin{itemize}
  \setlength\itemsep{1pt}
  \item \textbf{Diagnosis.} We show that the per-position marginals of a block diffusion drafter are dominated by its spine. After a fork, their top-1 accuracy, candidate lists and path probabilities all degrade, although the same drafter predicts these branches well once re-anchored.
  \item \textbf{Method.} We propose DForkTree, which re-anchors each branch on its own tokens without training. We derive from Bayes' rule why this evidence enters the score as PMI, show that its optimal weight follows what it adds beyond the drafter, and prove that the optimal tree under these scores can be built exactly on the GPU.
  \item \textbf{Evaluation.} So that the recurrent state of hybrid models does not bottleneck tree verification, we build a closed-form verifier for all tree methods. With the same drafter and GPU tree pipeline, DForkTree outperforms all baselines on two model sizes at temperatures 0 and 1. Our analyses trace nearly all of its extra accepted tokens to branches and show that neither calibrating DDTree by node type nor a trained selector matches it.
\end{itemize}

\section{Background}
\label{sec:background}

Let $p$ be the target model and $x_{\le t}$ the committed prefix.
In each round, a drafter proposes a tree $\mathcal{T}$ rooted at the last committed token $r$, and the target scores all of its nodes in one forward pass with an ancestor mask.
Under greedy decoding, the round accepts the longest path whose tokens match the target's choices and adds one bonus token.
We write $\tau$ for the mean number of tokens per round.

A block diffusion drafter such as DFlash reads the prefix once and predicts, in parallel, a marginal $q_k$ over the vocabulary for each of $K$ future positions.
A 24-token block thus holds the last committed token and $K=23$ draft positions.
DDTree scores a path $u=(u_1,\dots,u_d)$ by the product of its marginals and keeps the tree that would accept the most draft tokens in expectation under this product,
\begin{equation}
  Q(u)=\prod_{k=1}^{d} q_k(u_k), \quad \mathcal{T}^\star=\arg\max_{|\mathcal{T}|\le B} \sum_{u\in\mathcal{T}} Q(u).
  \label{eq:ddtree}
\end{equation}
Because $Q$ never increases along a path, $\mathcal{T}^\star$ consists of the $B$ prefixes with the largest $Q$, and a best-first heap finds it exactly.
We call the path of top-1 tokens $(\arg\max q_1,\arg\max q_2,\dots)$ the drafter's \emph{spine}.
A node \emph{forks} when its token is not the top-1 token at its position, and all of its descendants are then off the spine.

\section{Where the Marginals Fail}
\label{sec:analysis}

On held-out development prompts, we first follow the target's true continuation through the top-16 candidates of each position and record at every node whether the drafter's top-1 candidate is the true next token.
This yields about one million events, grouped by how each node relates to the spine.
A node is \emph{on the spine} if all its ancestors are top-1 tokens, \emph{just forked} if its parent is not top-1, and \emph{back on rank 1} if an earlier ancestor forked but its parent is again top-1.

\textbf{Accuracy falls at the fork.}
The groups differ sharply (\cref{fig:motivation}a).
The drafter's top-1 token is correct for 89\% of the spine nodes, which make up two thirds of the events, and for only 18\% of the nodes right after a fork.
Nodes that later return to a top-1 token recover to 63\%, and the deficit persists along the branch.
Each marginal averages over the prefixes that the drafter itself considers likely, and the spine dominates this average.
Once a path forks, the marginals at later positions are therefore still fitted mostly to the spine---a prefix that the branch no longer follows.
This likely also explains the partial recovery at rank 1.

\textbf{Candidates are lost too.}
The candidate lists come from the same marginals and show the same pattern.
In the trees of \cref{sec:sources}, the true next token is among the top 16 for 98\% of the spine nodes and for about three quarters of the nodes after a fork.
Prior work observes that coverage decays along the block and attributes this decline to the limited capacity of the drafter's backbone \citep{dflash2blog2026}.
\Cref{fig:motivation}c splits that measurement by whether the path is still on the spine.
Let $S_k$ and $F_k$ be the coverage on and off the spine at position $k$ and $p_k$ the share of paths still on the spine, with $p_1=1$.
The pooled coverage $R_k=p_kS_k+(1-p_k)F_k$ then falls from the first position to position $K$ by
\begin{equation}
  R_1-R_K=\underbrace{(S_1-S_K)}_{\text{on the spine}}+\underbrace{(1-p_K)(S_K-F_K)}_{\text{leaving the spine}}.
  \label{eq:decomp}
\end{equation}
Notably, coverage on the spine is still 98.5\% at the last of 23 positions, so the first term stays small.
The second term accounts for 95\% of the decline over the first seven positions for DFlash and for 90\% with DFlash 2 (Appendix~\ref{app:sens}).
Coverage thus decays mainly because deeper paths are more likely to have left the spine.

\textbf{Re-anchoring restores accuracy.}
If conditioning on the spine is indeed the cause, giving the drafter the true prefix of the branch should restore its accuracy.
For nodes right after the first fork on the target's true continuation, we re-anchor the drafter by re-running it from the fork token with the target's features of the prefix, as the next round would.
On these nodes, its top-1 accuracy on the following token rises from 29\% inside the original block to 84\%, and to 87\% and 89\% at the next two tokens (\cref{fig:motivation}b), and Qwen3.5-9B behaves almost identically.
The recovery depends on the target's features.
Filling the branch tokens into the drafter's block in place of the masks recovers little of it (\cref{sec:ceiling}), likely because DFlash is trained with masks after the first position, so such inputs lie outside its training distribution.

\textbf{Path probabilities are miscalibrated.}
The path probabilities that decide how a tree spends its budget suffer from the same assumption.
In DDTree's 32-node trees, nodes after a fork are accepted only 0.71 times as often as their path probabilities predict and spine nodes 1.20 times as often, with similar ratios for the EAGLE-3 tree.
Because all nodes compete for one budget, the tree moves nodes from the spine to branches that the scores overrate, and one global temperature cannot correct both groups at once.

Better scores after a fork should therefore help most, as long as they leave the accurate spine nearly untouched and allow for a true token outside the candidates.
The score defined in \cref{sec:method} (blue in \cref{fig:motivation}a) raises post-fork accuracy from 18\% to 50\% and barely changes the spine.

\section{DForkTree}
\label{sec:method}

\begin{figure*}[t]
  \centering
  \includegraphics[width=0.86\textwidth]{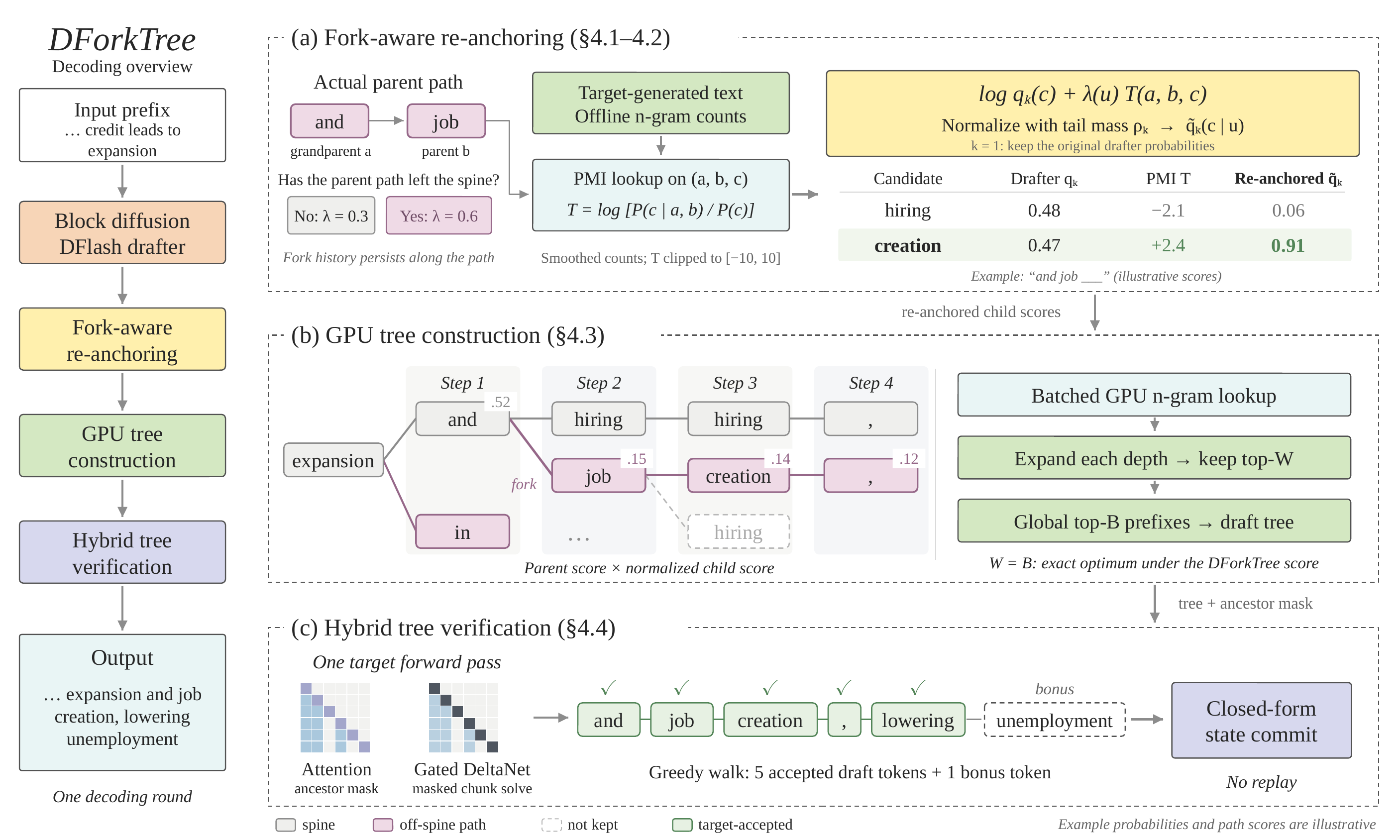}
  \vspace{-1.5mm}
  \caption{Overview of DForkTree on an MT-Bench example with illustrative numbers. (a) After the fork to \emph{job}, the marginals rate \emph{hiring} and \emph{creation} almost equally, and DForkTree re-scores each child by the PMI of its grandparent, parent and child, with a larger weight off the spine. (b) The tree grows one depth at a time on the GPU and keeps the best $B$ prefixes. (c) One target pass verifies the tree and commits the accepted path's state in closed form.}
  \label{fig:overview}
\end{figure*}

\begin{figure}[t]
  \centering
  \includegraphics[width=0.9\linewidth]{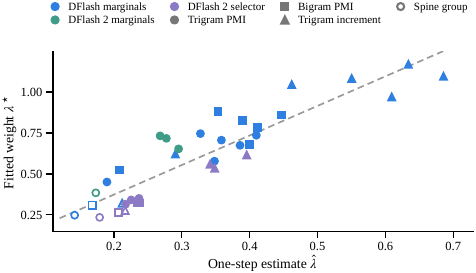}
  \vspace{-3.5mm}
  \caption{One-step estimate $\hat\lambda$ against the fitted weight $\lambda^\star$ for 37 groups of development nodes, fitted as $\lambda^\star\approx1.8\,\hat\lambda$ ($R^2=0.82$).}
  \label{fig:weighthat}
\end{figure}

\subsection{Path Statistics from Self-Generated Text}
\label{sec:stats}

Re-anchoring needs the target's features of each branch, which a single round does not provide, so DForkTree approximates it in the drafter's output space.
It leaves the drafter unchanged, modifies only how nodes are scored and how the tree is built (\cref{fig:overview}), and takes the target's local token dependencies from $n$-gram statistics of text that the target model generated itself, here 100K responses of Qwen3.5-4B to instruction prompts on math, code, chat and general requests.
We count unigrams, bigrams and trigrams once, offline, and smooth the counts with interpolated Dirichlet estimates \citep{mackay1995hierarchical}.
The target generates these 95.7M tokens once, counting them takes 27~s on one CPU thread, and the tables occupy 163~MB of GPU memory.
For a candidate $c$ that follows the tokens $(a,b)$ on a path, the statistic is the pointwise mutual information \citep{church1990word}
\begin{equation}
  T(a,b,c)=\log\frac{P(c\mid a,b)}{P(c)},
  \label{eq:pmi}
\end{equation}
where $P(c\mid a,b)$ and $P(c)$ are the trigram and unigram estimates and $T$ is clipped to $[-10,10]$.
$T$ is positive when the target generates $c$ after $(a,b)$ more often than in general.

\subsection{Fork-Aware Path Scores}
\label{sec:fork}

\begin{figure*}[t]
  \centering
  \includegraphics[width=0.495\textwidth]{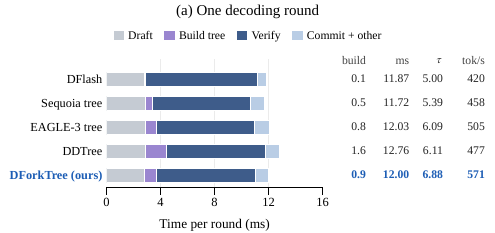}\hfill
  \includegraphics[width=0.495\textwidth]{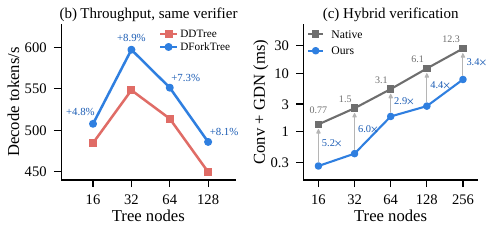}
  \vspace{-1mm}
  \caption{System cost on Qwen3.5-4B. (a) One decoding round for the 24-token DFlash chain and four tree methods with the budgets of \cref{tab:main}, split into four stages, with DDTree's host search counted as building. Columns give build and round time in ms, $\tau$ and throughput. (b) Throughput with 23 draft positions, labeled with DForkTree's gain over DDTree. (c) Short convolution and 24 GDN layers for one tree verification, with native snapshot sizes in GiB and our speedup.}
  \label{fig:system}
\end{figure*}

\begin{proposition}[Evidence combination]
\label{prop:poe}
Let $h$ be the information the drafter uses for a position and $(a,b)$ the two preceding tokens on a path.
If $h$ and $(a,b)$ are conditionally independent given the next token $c$, then
$p(c\mid h,a,b)\propto p(c\mid h)\,p(c\mid a,b)/p(c)$.
\end{proposition}

In log space, Proposition~\ref{prop:poe} corrects the drafter's log-probability by the PMI.
Adding the raw $n$-gram log-probability $\log P(c\mid a,b)$ instead, as DART does with external $n$-gram scores \citep{dart2026}, counts the unigram prior $p(c)$ twice, because the drafter's marginal $q_k(c)$ already reflects it.
Dividing by $p(c)$ removes this shared prior, as in density-ratio fusion of language models \citep{mcdermott2019density}.
Since the drafter's information $h$ and the path tokens $(a,b)$ are not exactly independent given $c$, we temper the correction with a weight $\lambda$, as in a product of experts.

Let $\mathcal{C}_k$ be the top-$M$ candidates of $q_k$ and $\rho_k=1-\sum_{z\in\mathcal{C}_k}q_k(z)$ the tail mass outside them, which we treat as one more class with neutral evidence $T=0$.
The first position keeps $\tilde q_1=q_1$.
For a path $u_{<k}$, with $u_0=r$ and $u_{-1}=x_{t-1}$, applying Proposition~\ref{prop:poe} to $\mathcal{C}_k$ and this tail class scores $c\in\mathcal{C}_k$ at position $k\ge 2$ by
\begin{equation}
  \tilde q_k(c\mid u_{<k})=\frac{q_k(c)\,e^{\lambda(u_{<k})\,T(u_{k-2},u_{k-1},c)}}{\rho_k+\sum_{z\in\mathcal{C}_k}q_k(z)\,e^{\lambda(u_{<k})\,T(u_{k-2},u_{k-1},z)}},
  \label{eq:score}
\end{equation}
and defines the path score recursively,
\begin{equation}
  \tilde Q(u_{\le k})=\tilde Q(u_{<k})\,\tilde q_k(u_k\mid u_{<k}).
  \label{eq:path}
\end{equation}
Keeping $\rho_k$ in the normalizer leaves room for a true token outside the candidates, which is common after a fork.
When every candidate has negative PMI with the branch prefix, their total score falls below the drafter's own $1-\rho_k$.
Renormalizing over the candidates, as UBTree \citep{ubtree2026} and the DFlash 2 selector that we test do, forces this total to one and makes branch scores overconfident (\cref{sec:ceiling}).
Through $T$, the score $\tilde Q$ depends on the tokens chosen along the path, which is how DForkTree re-anchors each branch on its own prefix.

At inference, the weight $\lambda(u_{<k})$ takes one of two constants, $\lambda_a=0.3$ if every token of $u_{<k}$ is the top-1 token at its position and $\lambda_p=0.6$ once the path has left the spine, for all benchmarks and both models.
After a fork, the $n$-grams condition on a prefix that $q_k$ does not reflect, so their evidence should count more than on the spine, where much of it is already in $q_k$.
Weights fitted by maximum likelihood for each group of nodes follow this order, at about 0.25 on the spine and at 0.6 or more after a fork (\cref{fig:lambdamain}).
Proposition~\ref{prop:weight}, which DForkTree does not use at inference, explains this order by how much the evidence overlaps with $q_k$.
For a group of nodes, let $L(\lambda)$ be the mean log-likelihood of the true next token $c^\star$ under \cref{eq:score}, with the tail mass as one more class with $T=0$.

\begin{proposition}[Weight of overlapping evidence]
\label{prop:weight}
$L$ is concave in $\lambda$, and its maximizer $\lambda^\star$ satisfies $\mathbb{E}[T(c^\star)]=\mathbb{E}\big[\mathbb{E}_{\tilde q_k}T\big]$.
One Newton step from $\lambda=0$ gives
\[
  \hat\lambda=\frac{\mathbb{E}\big[T(c^\star)-\mathbb{E}_{q_k}T\big]}{\mathbb{E}\big[\mathrm{Var}_{q_k}T\big]}.
\]
If $c^\star$ is distributed according to $q_k$ given everything that $T$ depends on, the numerator is zero and so is $\lambda^\star$.
\end{proposition}

Intuitively, the numerator of $\hat\lambda$ measures how strongly $T$ points to the true token beyond what $q_k$ already expects, so evidence that $q_k$ already reflects adds only to the denominator.
On the spine, the true token's PMI exceeds its expectation under $q_k$ by only 0.48, against about 4.5 after a fork, so $\hat\lambda$ rises from 0.14 to 0.33 or more.
Across 37 node groups from DFlash, DFlash 2 (\cref{sec:ceiling}) and three kinds of evidence (\cref{fig:weighthat}), $\hat\lambda$ ranks the fitted weights $\lambda^\star$ with a Spearman correlation of 0.89, and $\lambda^\star$ falls as the correlation between $\log q_k$ and $T$ over the candidates grows.

With $\lambda_a=\lambda_p=0$, DForkTree reduces to DDTree.

\subsection{Building the Optimal Tree Exactly}
\label{sec:build}

Although $\tilde Q$ depends on the path, the optimality result of DDTree carries over, and the optimal tree can be found on the GPU without a sequential heap.

\begin{proposition}[Optimal tree]
\label{prop:opt}
(a) If the continuation were drawn from $\tilde q$, a tree $\mathcal{T}$ would accept $\sum_{u\in\mathcal{T}}\tilde Q(u)$ draft tokens in expectation, and the $B$ prefixes with the largest $\tilde Q$ form a valid tree that maximizes this sum among all trees with $B$ nodes whose tokens at depth $k$ come from $\mathcal{C}_k$.
(b) Growing the tree depth by depth and keeping at each depth the $W$ children with the largest $\tilde Q$ among children of the kept nodes recovers exactly these $B$ prefixes if $W\ge B$.
\end{proposition}

Part (a) holds because $\tilde q_k\le1$, as in DDTree and Lemma~1 of DARTree \citep{dartree2026}, and part (b) follows from a counting argument (Appendix~\ref{app:proofs}).

In practice, DForkTree evaluates $T$ for all $(K-1)M^3$ combinations of grandparent, parent and child candidates in one batched lookup on the GPU.
Each depth of the search then takes a single top-$W$ with fixed shapes (\cref{alg:build}), so the whole construction compiles into a CUDA graph.
We set $W=B$ and do not force the spine into the tree.
Building takes 0.9~ms of a DForkTree round, compared with 0.8~ms for the EAGLE-3 tree and 1.6~ms for DDTree, whose heap runs on the host (\cref{fig:system}a).
Because a whole DForkTree round is only 1\% longer than a round of the 24-token chain, its longer accepted paths turn almost directly into the throughput gains of \cref{fig:system}b.

\begin{table*}[t]
\centering
\caption{Speedup $S$ over autoregressive decoding and acceptance length $\tau$ at temperature 0, best per model in \textbf{bold} and best baseline \underline{underlined}. MTP is Qwen3.5's built-in head with 4 steps and top-4, and the EAGLE-3 tree is EAGLE's dynamic tree on the DFlash drafter. All tree methods share the DFlash drafter, the GPU tree packing and our verifier and use 32 nodes, with blocks of 16 for the baselines and 24 for DForkTree.}
\label{tab:main}
\vspace{-1mm}
\setlength{\tabcolsep}{2.4pt}
\renewcommand{\arraystretch}{1.02}
\resizebox{\textwidth}{!}{%
\begin{tabular}{l >{\hspace{3.4pt}}c c<{\hspace{3.4pt}} >{\hspace{3.4pt}}c c<{\hspace{3.4pt}} >{\hspace{3.4pt}}c c<{\hspace{3.4pt}} >{\hspace{3.4pt}}c c<{\hspace{3.4pt}} >{\hspace{3.4pt}}c c<{\hspace{3.4pt}} >{\hspace{3.4pt}}c c<{\hspace{3.4pt}} >{\hspace{3.4pt}}c c<{\hspace{3.4pt}} >{\columncolor{avgbg}\hspace{3.4pt}}c >{\columncolor{avgbg}}c<{\hspace{3.4pt}}}
\toprule
{\boldmath$T=0$} & \multicolumn{2}{c}{GSM8K} & \multicolumn{2}{c}{MATH-500} & \multicolumn{2}{c}{HumanEval} & \multicolumn{2}{c}{MBPP} & \multicolumn{2}{c}{LCB} & \multicolumn{2}{c}{MT-Bench} & \multicolumn{2}{c}{Alpaca} & \multicolumn{2}{>{\columncolor{avgbg}}c}{Avg.} \\
\cmidrule(lr){2-3}\cmidrule(lr){4-5}\cmidrule(lr){6-7}\cmidrule(lr){8-9}\cmidrule(lr){10-11}\cmidrule(lr){12-13}\cmidrule(lr){14-15}\cmidrule(lr){16-17}
Method & $S$ & $\tau$ & $S$ & $\tau$ & $S$ & $\tau$ & $S$ & $\tau$ & $S$ & $\tau$ & $S$ & $\tau$ & $S$ & $\tau$ & $S$ & $\tau$ \\
\midrule
\rowcolor{modelbg}\multicolumn{17}{l}{\textbf{Qwen3.5-4B}} \\
Autoregressive & 1.00 & 1.00 & 1.00 & 1.00 & 1.00 & 1.00 & 1.00 & 1.00 & 1.00 & 1.00 & 1.00 & 1.00 & 1.00 & 1.00 & 1.00 & 1.00 \\
$N$-gram & 1.00 & 1.68 & 1.09 & 1.85 & 0.80 & 1.54 & 0.84 & 1.44 & 0.96 & 1.64 & 0.79 & 1.32 & 0.75 & 1.25 & 0.89 & 1.53 \\
MTP & 2.07 & 4.67 & 2.05 & 4.70 & 1.87 & 4.62 & 1.98 & 4.42 & 1.85 & 4.34 & 1.63 & 3.88 & 1.53 & 3.57 & 1.85 & 4.32 \\
DFlash, 16 tokens & 3.61 & 7.09 & 3.70 & 7.43 & 3.67 & 7.72 & 2.94 & 5.91 & 2.45 & 4.99 & 1.74 & 3.60 & 1.44 & 2.86 & 2.79 & 5.66 \\
DFlash, 24 tokens & 3.88 & 7.42 & 4.06 & 7.87 & 3.87 & 8.03 & 3.09 & 6.04 & 2.58 & 5.05 & 1.86 & 3.66 & 1.48 & 2.92 & 2.97 & 5.85 \\
\addlinespace[2pt]
Sequoia tree & 4.12 & 7.24 & 4.19 & 7.47 & 3.97 & 7.64 & 3.52 & 6.29 & 3.06 & 5.58 & 2.23 & 4.22 & 1.89 & 3.43 & 3.28 & 5.98 \\
DART tree & 3.36 & 7.62 & 3.59 & 8.24 & 3.13 & 7.75 & 3.10 & 6.77 & 2.59 & 5.92 & 1.99 & 4.62 & 1.69 & 3.81 & 2.78 & 6.39 \\
EAGLE-3 tree & 4.33 & 8.26 & \underline{4.76} & \underline{8.92} & 4.36 & 8.95 & \underline{3.99} & 7.29 & \underline{3.25} & 6.21 & \underline{2.46} & \underline{4.74} & \underline{2.07} & 3.84 & \underline{3.60} & 6.89 \\
DDTree & \underline{4.38} & \underline{8.39} & 4.56 & 8.87 & \underline{4.46} & \underline{9.12} & 3.73 & \underline{7.38} & 3.12 & \underline{6.25} & 2.41 & 4.72 & 1.99 & \underline{3.86} & 3.52 & \underline{6.94} \\
\addlinespace[2pt]
\rowcolor{oursbg}
\textbf{DForkTree} & \textbf{5.14} & \textbf{9.47} & \textbf{5.47} & \textbf{10.25} & \textbf{5.23} & \textbf{10.55} & \textbf{4.29} & \textbf{8.14} & \textbf{3.55} & \textbf{6.82} & \textbf{2.63} & \textbf{5.04} & \textbf{2.18} & \textbf{4.07} & \textbf{4.07} & \textbf{7.76} \\
\midrule
\rowcolor{modelbg}\multicolumn{17}{l}{\textbf{Qwen3.5-9B}} \\
Autoregressive & 1.00 & 1.00 & 1.00 & 1.00 & 1.00 & 1.00 & 1.00 & 1.00 & 1.00 & 1.00 & 1.00 & 1.00 & 1.00 & 1.00 & 1.00 & 1.00 \\
$N$-gram & 1.22 & 1.65 & 1.34 & 1.82 & 1.12 & 1.49 & 1.07 & 1.42 & 1.18 & 1.62 & 0.92 & 1.30 & 0.91 & 1.25 & 1.11 & 1.51 \\
MTP & 2.41 & 4.74 & 2.37 & 4.75 & 2.31 & 4.64 & 2.25 & 4.50 & 2.24 & 4.41 & 1.92 & 3.96 & 1.82 & 3.69 & 2.19 & 4.38 \\
DFlash, 16 tokens & 4.12 & 7.17 & 4.62 & 7.67 & 4.76 & 7.80 & 3.88 & 6.15 & 3.02 & 5.03 & 2.18 & 3.66 & 1.82 & 3.05 & 3.49 & 5.79 \\
DFlash, 24 tokens & 4.61 & 7.80 & 4.62 & 7.98 & 4.59 & 7.91 & 3.73 & 6.21 & 3.10 & 5.15 & 2.18 & 3.76 & 1.79 & 3.03 & 3.52 & 5.98 \\
\addlinespace[2pt]
Sequoia tree & 5.17 & 8.22 & 5.27 & 8.41 & 4.85 & 8.09 & 4.27 & 6.88 & 3.59 & 5.84 & 2.61 & 4.28 & 2.19 & 3.61 & 3.99 & 6.48 \\
DART tree & 4.01 & 7.98 & 4.44 & 8.40 & 4.52 & 8.22 & 3.80 & 6.94 & 3.19 & 6.00 & 2.48 & 4.73 & 2.14 & 4.00 & 3.51 & 6.61 \\
EAGLE-3 tree & 5.45 & 9.01 & \underline{5.65} & 9.22 & \underline{5.78} & \underline{9.25} & \underline{4.66} & 7.60 & \underline{3.91} & 6.42 & \underline{2.95} & 4.92 & \underline{2.45} & \underline{4.10} & \underline{4.41} & 7.22 \\
DDTree & \underline{5.58} & \underline{9.20} & 5.43 & \underline{9.26} & 5.25 & 9.23 & 4.57 & \underline{7.63} & 3.82 & \underline{6.43} & 2.84 & \underline{4.93} & 2.40 & 4.09 & 4.27 & \underline{7.25} \\
\addlinespace[2pt]
\rowcolor{oursbg}
\textbf{DForkTree} & \textbf{5.92} & \textbf{9.91} & \textbf{6.19} & \textbf{10.40} & \textbf{5.91} & \textbf{10.43} & \textbf{5.23} & \textbf{8.35} & \textbf{4.34} & \textbf{7.04} & \textbf{3.06} & \textbf{5.25} & \textbf{2.62} & \textbf{4.34} & \textbf{4.75} & \textbf{7.96} \\
\bottomrule
\end{tabular}}
\end{table*}

\subsection{Verifying Trees on Hybrid Models}
\label{sec:verify}

Qwen3.5 interleaves attention with Gated DeltaNet (GDN) layers \citep{yang2025gated}, whose matrix-valued recurrent state differs from path to path in a tree.
SGLang's native tree verification \citep{zheng2024sglang} therefore writes a snapshot of this state for every node, up to 12.3~GiB per round at 256 nodes, and replays the accepted path to commit it.
So that this cost does not mask the differences between tree methods, we build a verifier without snapshots and use it for every tree method.
It follows the ancestor-masked chunk form of \citet{oda2026weaver}, so that each node keeps only its key $k_t$, its corrected value $\tilde v_t$ from the triangular solve and its cumulative log-decay $G_t$ along the path.
Once the target has accepted a path, unrolling the gated delta rule along it gives the new state in closed form from these quantities (Appendix~\ref{app:impl}), and a single fused kernel evaluates it for all GDN layers, heads and accepted nodes, so no state is stored per node, replayed or copied to the host.
The GDN layers and their short convolutions then run up to 6.0$\times$ faster than with native verification (\cref{fig:system}c), and every tree baseline becomes up to 17\% faster end to end (Appendix~\ref{app:system}).

\section{Experiments}
\label{sec:exp}

\subsection{Setup}
\label{sec:setup}

\paragraph{Models and serving.}
We evaluate Qwen3.5-4B and 9B \citep{qwen35} with their released BF16 DFlash drafters in SGLang 0.5.22 on RTX 5090 GPUs, at batch size one with thinking disabled.

\paragraph{Benchmarks and metrics.}
Following DFlash and DDTree, we take 128 prompts each from GSM8K \citep{cobbe2021gsm8k}, MATH-500 \citep{hendrycks2021math,lightman2024verify}, HumanEval \citep{chen2021humaneval}, MBPP \citep{austin2021mbpp}, LiveCodeBench (LCB) \citep{jain2025livecodebench} and Alpaca \citep{li2023alpacaeval} and add the first turns of the 80 MT-Bench prompts \citep{zheng2023judging}, for 848 prompts with at most 2048 new tokens each.
The speedup $S$ divides the decoding time of autoregressive decoding by that of the method, and $\tau$ counts output tokens per round, including the bonus token.
Because outputs differ in length between methods, the throughput margins in the text pool tokens and time over all 848 prompts and differ from ratios of the average $S$.

\paragraph{Baselines.}
We compare with SGLang's $n$-gram lookup, the MTP head of Qwen3.5 and DFlash chains of 16 and 24 tokens.
For trees, we pair the same DFlash drafter with Sequoia's static tree \citep{chen2024sequoia}, DART's $n$-gram tree with its official C++ search \citep{dart2026}, the dynamic tree of EAGLE-2 and EAGLE-3 \citep{li2024eagle2,li2025eagle3}, called the EAGLE-3 tree, and DDTree with its official heap \citep{ringel2026ddtree}.
All trees share our verifier and GPU packing, keep their own construction and defaults, and use 32 nodes with their best block size, 16 for the baselines and 24 for DForkTree (Appendix~\ref{app:baselines}).
At the same 23 draft positions, DForkTree remains faster than DDTree at every budget (\cref{fig:system}b).
DForkTree uses $M=16$, and Qwen3.5-9B reuses the counts and weights of Qwen3.5-4B.

\subsection{Main Results}
\label{sec:main}

DForkTree has the longest acceptance and the highest speedup on every benchmark for both models (\cref{tab:main}).
\begin{table}[t]
\centering
\caption{Ablations on Qwen3.5-4B with 64 nodes. Gray settings are shared with DForkTree, row (e). $K$ counts draft positions, $\tau$ is averaged over benchmarks, and $\Delta$ is the throughput change.}
\label{tab:ablation}
\vspace{-2mm}
\small
\setlength{\tabcolsep}{2.3pt}
\begin{tabular}{l ll cc ccc}
\toprule
 & \multicolumn{2}{c}{Path score} & \multicolumn{2}{c}{Tree} & & & \\
\cmidrule(lr){2-3}\cmidrule(lr){4-5}
 & Evidence & Weight & Spine & $K$ & $\tau$ & tok/s & $\Delta$ \\
\midrule
(a) & none & -- & \dm{free} & \dm{23} & 7.72 & 515 & $-7.0\%$ \\
(b) & raw log-prob & single $\lambda$ & \dm{free} & \dm{23} & 8.00 & 531 & $-4.2\%$ \\
(c) & bigram PMI & single $\lambda$ & \dm{free} & \dm{23} & 8.17 & 543 & $-2.1\%$ \\
(d) & \dm{trigram PMI} & single $\lambda$ & \dm{free} & \dm{23} & 8.18 & 550 & $-0.7\%$ \\
\rowcolor{oursbg} \textbf{(e)} & \textbf{trigram PMI} & \textbf{fork-aware} & \textbf{free} & \textbf{23} & \textbf{8.36} & \textbf{554} & -- \\
\addlinespace[2pt]
(f) & \dm{trigram PMI} & \dm{fork-aware} & forced & \dm{23} & 8.17 & 543 & $-2.0\%$ \\
(g) & \dm{trigram PMI} & \dm{fork-aware} & \dm{free} & 15 & 7.86 & 536 & $-3.2\%$ \\
\bottomrule
\end{tabular}
\vspace{-1mm}
\end{table}
On Qwen3.5-4B, it raises throughput by 9.9\% over the strongest tree baseline, the EAGLE-3 tree, and by 15.1\% over DDTree, and on Qwen3.5-9B by 8.1\% and 11.0\%.
On average, it decodes $4.07\times$ and $4.75\times$ faster than autoregressive decoding and about 41\% faster than the 24-token DFlash chain.
Unlike the tree baselines, the chain keeps SGLang's native verification, which takes 8.3 against 7.3~ms per round (\cref{fig:system}a), so the margin over the chain also includes our verifier.
According to Appendix~\ref{app:quality}, DForkTree's outputs match autoregressive decoding about as often as native DFlash does.

\subsection{Ablations}
\label{sec:ablation}

\Cref{tab:ablation} varies the components of DForkTree at 64 nodes, where branches hold a larger part of the tree.
Row (a) scores nodes by the marginals alone but keeps our depth, budget and construction, so it isolates the path prior, which adds 7.6\% in throughput.
Most of this gain comes from the bigram PMI in row (c), a training-free counterpart of first-order edge scorers such as GRAFT \citep{graft2026}.
The trigram in row (d) and the fork-aware weight in row (e) raise $\tau$ further to 8.36.
Consistent with Proposition~\ref{prop:poe}, adding the raw $n$-gram log-probability as DART does (row (b)) counts the token frequency twice, and the PMI of row (d) is 3.6\% faster.
Drafting only 15 positions costs 3.2\% (row (g)), so part of DForkTree's lead in \cref{tab:main} comes from the longer block that its branch scores make worthwhile, and forcing the spine into the tree costs 2.0\% (row (f)).

\subsection{Where the Extra Tokens Come From}
\label{sec:sources}

\begin{figure}[t]
  \centering
  \includegraphics[width=\linewidth]{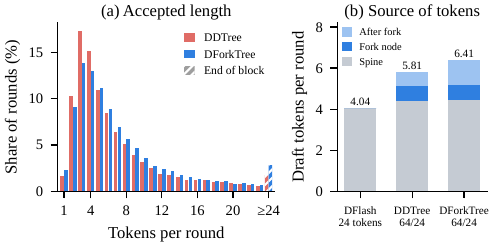}
  \vspace{0mm}
  \caption{Replay of the 24-token DFlash chain and of DDTree and DForkTree with 64 nodes and 23 draft positions. (a) Tokens produced per round. A round produces at most 24 tokens, so the hatched bar counts all rounds that reach the end of the block. (b) Accepted draft tokens per round on the spine, at the fork node where the accepted path leaves it, and after the fork.}
  \label{fig:sources}
  \vspace{2mm}
\end{figure}

To check the diagnosis on whole rounds, we replay the 24-token DFlash chain, DDTree and DForkTree with 64 nodes, which reproduces the online $\tau$ within 1\%, and classify each accepted draft token by its position relative to the spine.
DForkTree accepts 0.60 more draft tokens per round than DDTree, and 96\% of them lie at or after the fork (\cref{fig:sources}b), mainly in the long rounds (\cref{fig:sources}a).
Both trees spend about 55 of their 63 nodes off the spine (\cref{tab:shape}), and DForkTree's branches reach deeper (mean maximum depth 10.3 vs.\ 8.3), with 38\% more accepted tokens per off-spine node.

After a fork, DForkTree's nodes are accepted 0.94 times as often as its path scores predict, against 0.71 for DDTree.
Temperatures fitted per node type bring DDTree's ratio close to one but raise its offline $\tau$ by only 1.1\%, far below DForkTree's lead of 8.0\%, so better calibration accounts for little of the gain.

\subsection{Comparison with Learned Corrections}
\label{sec:ceiling}

\textbf{Accuracy after a fork.}
\Cref{tab:ceiling} measures each correction by the share of the top-1 accuracy gap to its own re-anchored drafter that it closes at the first three tokens after a fork.
Filling the branch tokens into the DFlash drafter closes at most 16\%.
We also test two learned scorers, a small head distilled from the re-anchored DFlash drafter that reads the parent and grandparent (Appendix~\ref{app:sens}), and the trained selector of DFlash 2 \citep{dflash2blog2026}, which scores each candidate given its parent.
Within each row, DForkTree's PMI closes nearly as much of the gap as the selector and more than the distilled head, and adding it to either scorer brings the closure to about half.
Up to a fifth of the gap lies outside the 16 candidates that all of these scores re-rank.

\textbf{What counts add to a selector.}
Because the selector already conditions on the parent, the bigram part $\log P(c\mid b)/P(c)$ of the PMI raises the closed share by at most 1.7 points with a fitted weight of 0.15--0.20.
The trigram increment $\log P(c\mid a,b)/P(c\mid b)$ carries the grandparent, which the selector does not see, and adds about 4 points with a weight near 0.5, as Proposition~\ref{prop:weight} predicts.

\textbf{Full trees on DFlash 2.}
In full trees the order reverses, and the selector accepts 1.9\% fewer tokens than DForkTree.
The selector renormalizes over its 16 candidates and drops the tail mass $\rho_k$, so at 32 nodes it predicts an acceptance rate of 8.7\% for branch nodes that are accepted 5.4\% of the time, and restoring $\rho_k$ recovers a third of the gap.
Restoring the drafter's marginals at the children of spine nodes lowers the selector's $\tau$ by 5.7\%, so most of its gain comes from re-ranking at these forks, where DForkTree's PMI does at least as well.

\textbf{Upper bound.}
An oracle that re-anchors every kept branch node up to three tokens after a fork gains another 10.4\% and 13.0\% in offline $\tau$ at 32 and 64 nodes, and DForkTree recovers about 40\% of this gap without training.
A second, LoRA-tuned drafter pass over the branch adds 2.5 to 7.7~ms per round for at most 2.9\% higher offline $\tau$ (Appendix~\ref{app:sens}).

\subsection{Node Budget, Corpus and Sampling}
\label{sec:budget}
\label{sec:corpus}
\label{sec:sampling}

\paragraph{Node budget.}
DForkTree and DDTree are both fastest at 32 nodes (\cref{fig:system}b).
\begin{table}[t]
\centering
\caption{Share of the gap to the re-anchored drafter closed at the first three tokens after a fork (range, \%). Rows use different events, so only numbers within a row compare.}
\label{tab:ceiling}
\vspace{-2mm}
\small
\setlength{\tabcolsep}{0pt}
\begin{tabular*}{\linewidth}{@{\extracolsep{\fill}} l l c c >{\columncolor{oursbg}[3pt][3pt]}c c @{}}
\toprule
Correction & Drafter & Trained & Alone & DForkTree & Both \\
\midrule
Branch tokens & DFlash & & 11--16 & 41--46 & -- \\
Distilled head & DFlash & \cmark & 34--38 & 36--45 & 41--49 \\
Selector & DFlash 2 & \cmark & 46--47 & 44--45 & 49--50 \\
\bottomrule
\end{tabular*}
\end{table}
Offline, DForkTree has the highest $\tau$ from 8 to 256 nodes, leading DDTree by up to 9\% (\cref{fig:analysis}).

\paragraph{Corpus.}
With only 1K responses, DForkTree already reaches 532 tokens/s against 514 for DDTree, and offline $\tau$ grows roughly linearly in the logarithm of the corpus size up to 100K responses (Appendix~\ref{app:corpus}).
On the same 10K math and code prompts, the target's own answers give a higher $\tau$ than published reference solutions, likely because they follow the target's phrasing, although they also contain 2.7 times as many tokens.
Removing all math or all code lowers $\tau$ on the held-out domain by only 2--3\%.

\suppressfloats[t]
\begin{table}[t]
\centering
\caption{Averages at temperature 1, with $S$ the throughput ratio to autoregressive sampling. Appendix~\ref{app:t1} gives all methods per benchmark.}
\label{tab:t1}
\vspace{-2mm}
\small
\setlength{\tabcolsep}{3pt}
\begin{tabular}{l ccc ccc}
\toprule
& \multicolumn{3}{c}{Qwen3.5-4B} & \multicolumn{3}{c}{Qwen3.5-9B} \\
\cmidrule(lr){2-4}\cmidrule(lr){5-7}
Method & $S$ & $\tau$ & tok/s & $S$ & $\tau$ & tok/s \\
\midrule
MTP & 1.69 & 3.90 & 282 & 2.03 & 4.04 & 194 \\
DFlash, 16 tokens & 2.15 & 4.36 & 325 & 2.88 & 4.76 & 243 \\
DFlash, 24 tokens & 2.31 & 4.53 & 341 & 2.91 & 4.97 & 241 \\
EAGLE-3 tree & 2.80 & 5.69 & 429 & 3.64 & 6.23 & 310 \\
DDTree & 2.61 & 5.64 & 404 & 3.44 & 6.16 & 297 \\
\rowcolor{oursbg} \textbf{DForkTree} & \textbf{3.05} & \textbf{6.17} & \textbf{465} & \textbf{3.88} & \textbf{6.73} & \textbf{331} \\
\bottomrule
\end{tabular}
\end{table}

\paragraph{Sampling.}
At temperature 1, all tree methods use the lossless sample-and-match verification of DDTree.
DForkTree keeps its lead on both models (\cref{tab:t1}), with 8.5\% and 6.7\% higher throughput than the EAGLE-3 tree.
In-budget accuracy stays on par with autoregressive sampling, and token-level replays confirm that sampling is lossless (Appendix~\ref{app:t1}).

\section{Related Work}
\label{sec:related}

\paragraph{Speculative decoding and draft trees.}
Tree verification \citep{miao2024specinfer} checks many continuations in one pass.
Medusa \citep{cai2024medusa}, EAGLE-2 and EAGLE-3 \citep{li2024eagle2,li2025eagle3} draft with heads on the target, and Sequoia \citep{chen2024sequoia} optimizes static trees.
Prompt lookup \citep{saxena2023pld} and REST \citep{he2024rest} draft directly from $n$-grams.

\paragraph{Trees from block diffusion drafters.}
Follow-up work on DFlash \citep{dflash2026} and DDTree \citep{ringel2026ddtree} restores the dependencies between positions with trained selectors and edge scorers \citep{ubtree2026,graft2026,dflash2blog2026}, sequential modules \citep{dartree2026,dspark2026}, a pairwise corrector \citep{eisenstadt2026dcoco} or the adapter Weaver \citep{oda2026weaver}, which lists mid-draft re-anchoring as future work.
Carryover drafting \citep{carryover2026} and DFlow \citep{dflow2026} pass the target's hidden states of rejected tokens to the next round, consistent with our re-anchoring results.
DominoTree \citep{dominotree2026}, TreeSpark \citep{treespark2026} and CAST \citep{cast2026} report miscalibrated scores, and we find that the sign of the error depends on whether a node is on the spine.
DART \citep{dart2026} and PRESTO \citep{wang2026presto} add $n$-gram scores with one weight for all nodes.

\paragraph{Linear attention and tree verification.}
STree \citep{wu2025stree}, Weaver, DominoTree and UBTree verify trees on hybrid recurrent models \citep{gu2023mamba,yang2025gated}, and our verifier adds a closed-form state commit.

\section{Conclusion}
\label{sec:conclusion}

Block diffusion draft trees gain over chains on branches that leave the drafter's spine, and these are the nodes where its marginals, candidates and path probabilities break down.
Re-anchored at the fork, the same drafter continues these branches well, which points to the assumed spine prefix as the cause.
DForkTree re-scores branches without training using PMI from target-generated $n$-grams, weighted more heavily after a fork, builds the exact optimal tree on the GPU and is the fastest method we compare.
On DFlash 2, our counts beat a trained selector in full trees, so this mismatch is not specific to one drafter.
Since counted $n$-grams and learned corrections close about half of the gap to a re-anchored drafter, closing the rest likely requires the target's own representation of the branch.
Drafters that condition on the branch they extend are a natural next step.

\vfill
\clearpage
\bibliography{refs}
\bibliographystyle{icml2026}

\newpage
\appendix
\raggedbottom
\onecolumn

\section{Results at Temperature 1}
\label{app:t1}

\begin{table}[!h]
\centering
\caption{Speedup $S$ and acceptance length $\tau$ at temperature 1, with top-p 1 and no top-k (848 prompts per model). Sampled outputs differ between methods, so $S$ is the ratio of decoding throughputs to autoregressive sampling on each benchmark. Autoregressive sampling runs at 169 and 97 tok/s on Qwen3.5-4B and 9B, DForkTree is best in every column, and the best baseline is \underline{underlined}.}
\label{tab:t1full}
\setlength{\tabcolsep}{2.4pt}
\renewcommand{\arraystretch}{1.02}
\resizebox{\textwidth}{!}{%
\begin{tabular}{l >{\hspace{3.4pt}}c c<{\hspace{3.4pt}} >{\hspace{3.4pt}}c c<{\hspace{3.4pt}} >{\hspace{3.4pt}}c c<{\hspace{3.4pt}} >{\hspace{3.4pt}}c c<{\hspace{3.4pt}} >{\hspace{3.4pt}}c c<{\hspace{3.4pt}} >{\hspace{3.4pt}}c c<{\hspace{3.4pt}} >{\hspace{3.4pt}}c c<{\hspace{3.4pt}} >{\columncolor{avgbg}\hspace{3.4pt}}c >{\columncolor{avgbg}}c<{\hspace{3.4pt}}}
\toprule
{\boldmath$T=1$} & \multicolumn{2}{c}{GSM8K} & \multicolumn{2}{c}{MATH-500} & \multicolumn{2}{c}{HumanEval} & \multicolumn{2}{c}{MBPP} & \multicolumn{2}{c}{LCB} & \multicolumn{2}{c}{MT-Bench} & \multicolumn{2}{c}{Alpaca} & \multicolumn{2}{>{\columncolor{avgbg}}c}{Avg.} \\
\cmidrule(lr){2-3}\cmidrule(lr){4-5}\cmidrule(lr){6-7}\cmidrule(lr){8-9}\cmidrule(lr){10-11}\cmidrule(lr){12-13}\cmidrule(lr){14-15}\cmidrule(lr){16-17}
Method & $S$ & $\tau$ & $S$ & $\tau$ & $S$ & $\tau$ & $S$ & $\tau$ & $S$ & $\tau$ & $S$ & $\tau$ & $S$ & $\tau$ & $S$ & $\tau$ \\
\midrule
\rowcolor{modelbg}\multicolumn{17}{l}{\textbf{Qwen3.5-4B}} \\
Autoregressive & 1.00 & 1.00 & 1.00 & 1.00 & 1.00 & 1.00 & 1.00 & 1.00 & 1.00 & 1.00 & 1.00 & 1.00 & 1.00 & 1.00 & 1.00 & 1.00 \\
$N$-gram & 0.94 & 1.62 & 1.04 & 1.78 & 0.88 & 1.52 & 0.82 & 1.41 & 0.90 & 1.53 & 0.77 & 1.33 & 0.73 & 1.25 & 0.87 & 1.49 \\
MTP & 1.90 & 4.40 & 1.95 & 4.48 & 1.80 & 4.15 & 1.76 & 4.06 & 1.68 & 3.83 & 1.48 & 3.40 & 1.28 & 2.96 & 1.69 & 3.90 \\
DFlash, 16 tokens & 2.85 & 5.78 & 3.05 & 6.17 & 2.59 & 5.26 & 2.17 & 4.39 & 1.81 & 3.64 & 1.45 & 2.94 & 1.16 & 2.36 & 2.15 & 4.36 \\
DFlash, 24 tokens & 3.02 & 5.96 & 3.26 & 6.40 & 3.02 & 5.94 & 2.22 & 4.36 & 1.94 & 3.76 & 1.52 & 2.99 & 1.18 & 2.32 & 2.31 & 4.53 \\
Sequoia tree & 3.15 & 6.25 & 3.35 & 6.61 & 3.03 & 5.97 & 2.52 & 4.96 & 2.22 & 4.35 & 1.80 & 3.56 & 1.43 & 2.84 & 2.50 & 4.94 \\
DART tree & 2.72 & 6.74 & 2.88 & 7.09 & 2.53 & 6.27 & 2.22 & 5.51 & 1.97 & 4.80 & 1.59 & 3.93 & 1.28 & 3.16 & 2.17 & 5.36 \\
EAGLE-3 tree & \underline{3.63} & \underline{7.42} & \underline{3.77} & \underline{7.66} & \underline{3.41} & \underline{6.96} & 2.77 & 5.65 & \underline{2.46} & \underline{4.96} & \underline{1.97} & 4.01 & \underline{1.56} & \underline{3.18} & \underline{2.80} & \underline{5.69} \\
DDTree & 3.21 & 6.94 & 3.54 & 7.60 & 3.13 & 6.78 & \underline{2.80} & \underline{6.07} & 2.28 & 4.87 & 1.87 & \underline{4.04} & 1.47 & 3.17 & 2.61 & 5.64 \\
\rowcolor{oursbg}
\textbf{DForkTree} & \textbf{3.76} & \textbf{7.66} & \textbf{4.29} & \textbf{8.67} & \textbf{3.84} & \textbf{7.79} & \textbf{3.11} & \textbf{6.31} & \textbf{2.66} & \textbf{5.34} & \textbf{2.03} & \textbf{4.12} & \textbf{1.64} & \textbf{3.32} & \textbf{3.05} & \textbf{6.17} \\
\midrule
\rowcolor{modelbg}\multicolumn{17}{l}{\textbf{Qwen3.5-9B}} \\
Autoregressive & 1.00 & 1.00 & 1.00 & 1.00 & 1.00 & 1.00 & 1.00 & 1.00 & 1.00 & 1.00 & 1.00 & 1.00 & 1.00 & 1.00 & 1.00 & 1.00 \\
$N$-gram & 1.17 & 1.61 & 1.32 & 1.80 & 1.09 & 1.49 & 1.04 & 1.42 & 1.11 & 1.51 & 0.94 & 1.28 & 0.88 & 1.21 & 1.08 & 1.48 \\
MTP & 2.28 & 4.56 & 2.31 & 4.60 & 2.15 & 4.31 & 2.09 & 4.17 & 1.98 & 3.93 & 1.77 & 3.52 & 1.60 & 3.20 & 2.03 & 4.04 \\
DFlash, 16 tokens & 3.85 & 6.38 & 4.03 & 6.64 & 3.58 & 5.93 & 2.97 & 4.92 & 2.34 & 3.85 & 1.89 & 3.13 & 1.51 & 2.49 & 2.88 & 4.76 \\
DFlash, 24 tokens & 3.67 & 6.28 & 4.11 & 7.01 & 3.95 & 6.77 & 2.95 & 5.04 & 2.35 & 3.99 & 1.85 & 3.16 & 1.47 & 2.51 & 2.91 & 4.97 \\
Sequoia tree & 4.22 & 7.15 & 4.46 & 7.49 & 4.05 & 6.85 & 3.34 & 5.65 & 2.77 & 4.65 & 2.19 & 3.70 & 1.78 & 3.00 & 3.26 & 5.50 \\
DART tree & 3.71 & 7.41 & 3.89 & 7.70 & 3.48 & 6.93 & 3.02 & 6.04 & 2.53 & 4.99 & 2.08 & 4.13 & 1.71 & 3.41 & 2.92 & 5.80 \\
EAGLE-3 tree & \underline{4.79} & \underline{8.24} & \underline{4.87} & \underline{8.34} & \underline{4.50} & \underline{7.74} & \underline{3.83} & \underline{6.57} & \underline{3.01} & 5.12 & \underline{2.46} & 4.22 & \underline{1.99} & 3.41 & \underline{3.64} & \underline{6.23} \\
DDTree & 4.38 & 7.87 & 4.61 & 8.25 & 4.27 & 7.67 & 3.59 & 6.44 & 2.88 & \underline{5.14} & 2.39 & \underline{4.28} & 1.95 & \underline{3.49} & 3.44 & 6.16 \\
\rowcolor{oursbg}
\textbf{DForkTree} & \textbf{5.11} & \textbf{8.89} & \textbf{5.32} & \textbf{9.23} & \textbf{4.81} & \textbf{8.39} & \textbf{3.96} & \textbf{6.89} & \textbf{3.25} & \textbf{5.61} & \textbf{2.61} & \textbf{4.54} & \textbf{2.08} & \textbf{3.60} & \textbf{3.88} & \textbf{6.73} \\
\bottomrule
\end{tabular}}
\end{table}

\Cref{tab:t1full} gives the full results at temperature 1 under the protocol of \cref{tab:main}, with one sampled output per prompt and method.
Tree methods use sample-and-match verification, while the chains and MTP use the native sampling rule of SGLang, and Sequoia keeps the tree it calibrated at temperature 0.
On Qwen3.5-4B, DForkTree's throughput is 8.5\% higher than that of the EAGLE-3 tree (95\% CI 7.1--10.0\%), 15.2\% higher than that of DDTree (13.7--16.7\%) and 36.3\% higher than that of the 24-token DFlash chain (34.3--38.4\%).
On Qwen3.5-9B, the three margins are 6.7\% (5.3--8.1\%), 11.4\% (10.2--12.9\%) and 37.4\% (35.6--39.0\%).
The in-budget accuracy of the three tree methods on GSM8K, MATH-500, HumanEval and MBPP differs from autoregressive sampling by at most 6.3 points, and 23 of the 24 95\% intervals include zero.
The one exception is DForkTree on MBPP with Qwen3.5-9B.
With 24 comparisons and one sampled output per prompt, chance alone predicts about one such case, and the sampling checks below find no difference in the output distribution.
\Cref{tab:t1} in \cref{sec:sampling} gives the averages.

\paragraph{Sampling check.}
For every tree method and model, we replayed three generations of up to 512 tokens and checked that each committed token falls into the target's sampling interval at its node.
All 1,538 and 1,381 committed tokens of DForkTree pass on the two models, and the only miss among all methods is one token of the EAGLE-3 tree on Qwen3.5-4B that lies on a floating-point boundary of its interval.
The uniform numbers behind the committed tokens pass a Kolmogorov--Smirnov test for uniformity for every method ($p\ge0.25$).
On 10 prompts we also drew 500 short samples each from autoregressive sampling, DDTree and DForkTree.
Position-wise $\chi^2$ tests over the first eight output tokens find no difference, with $p>0.01$ in all 80 tests per method and model and a smallest $p$ of 0.048 for DForkTree.

\FloatBarrier
\section{Output Agreement and In-Budget Accuracy}
\label{app:quality}

\Cref{tab:quality} reports outputs identical to autoregressive decoding and in-budget accuracy under the 2048-token limit.
GSM8K uses the last number and MATH-500 the last complete boxed answer, both after text normalization.
HumanEval and MBPP run the original unit tests in a sandbox, and all 256 reference solutions pass them.
MBPP prompts often do not name the function, so all methods pass fewer than 10\% of MBPP tests as written.
We therefore report the rate with a fixed alias when the output defines a single top-level function.
Differences from autoregressive decoding stay within six problems per benchmark, and within four for the tree methods, and go in both directions.
DForkTree, for example, solves four more MATH-500 problems on Qwen3.5-4B and three fewer GSM8K and MATH-500 problems on Qwen3.5-9B.
About a third of the outputs of every speculative method match autoregressive decoding token by token, and native DFlash without a tree is no exception.
Multi-token verification changes the order of BF16 reductions and can flip near-ties in the greedy choice.

\begin{table}[h]
\centering
\small
\setlength{\tabcolsep}{4pt}
\fittab{Outputs identical to autoregressive decoding (out of 848) and in-budget accuracy (\%).\label{tab:quality}}{%
\begin{tabular}{l ccccc ccccc}
\toprule
& \multicolumn{5}{c}{Qwen3.5-4B} & \multicolumn{5}{c}{Qwen3.5-9B} \\
\cmidrule(lr){2-6}\cmidrule(lr){7-11}
Method & identical & GSM8K & MATH & HumanEval & MBPP & identical & GSM8K & MATH & HumanEval & MBPP \\
\midrule
Autoregressive & 848 & 91.4 & 68.8 & 84.4 & 65.6 & 848 & 94.5 & 74.2 & 88.3 & 68.0 \\
DFlash, 16 tokens & 286 & 91.4 & 70.3 & 85.9 & 67.2 & 316 & 93.0 & 74.2 & 88.3 & 66.4 \\
DFlash, 24 tokens & 276 & 90.6 & 73.4 & 83.6 & 67.2 & 298 & 93.0 & 72.7 & 89.8 & 67.2 \\
EAGLE-3 tree & 272 & 90.6 & 69.5 & 82.8 & 66.4 & 302 & 92.2 & 71.9 & 88.3 & 67.2 \\
DDTree & 282 & 92.2 & 68.8 & 85.2 & 68.0 & 304 & 93.0 & 74.2 & 86.7 & 66.4 \\
DForkTree & 281 & 91.4 & 71.9 & 83.6 & 65.6 & 306 & 92.2 & 71.9 & 87.5 & 67.2 \\
\bottomrule
\end{tabular}}
\end{table}

\paragraph{Repeated runs.}
We ran DFlash with 24 tokens and DForkTree with 64 nodes three times each on A210.
DFlash reached 434.3, 437.1 and 436.7 tokens/s and DForkTree 554.8, 557.1 and 557.3 tokens/s, a coefficient of variation of 0.35\% and 0.25\%.

\paragraph{Prompt sets.}
The test set has 848 prompts, 128 from each of GSM8K, MATH-500, HumanEval, MBPP, LiveCodeBench and Alpaca and the first turns of the 80 MT-Bench prompts.
A210 is a fixed subset of 210 test prompts on which we run the online analyses and replays of \cref{sec:analysis,sec:sources,sec:ceiling,sec:budget}.
The development analyses use 155 trajectories with blocks of 16 tokens, which give 1.03M events in \cref{fig:motivation}a, and \cref{fig:motivation}b uses 20K, 5K and 5K nodes at the first three tokens after a fork.
The weights and the smoothing are selected by two-fold cross-validation on the development replays.
\Cref{fig:system}a averages 42 test prompts on one GPU, with CUDA events that add 0.7--2.1\% to a round, and each point of \cref{fig:system}c averages 50 trees.
The oracle of \cref{sec:ceiling} runs on 28 prompts of A210, where it gains 10.4\% and 13.0\% in offline $\tau$ at 32 and 64 nodes and DForkTree recovers 39--41\% of the gap to it.

\paragraph{Confidence intervals.}
All intervals are paired-bootstrap 95\% CIs over prompts.
At temperature 0, DForkTree's throughput margins over the EAGLE-3 tree and DDTree in \cref{sec:main} have intervals of 8.9--10.9\% and 14.0--16.2\% on Qwen3.5-4B and of 7.1--9.0\% and 10.0--12.0\% on Qwen3.5-9B.
In \cref{tab:ablation}, the gain of the trigram in row (d) over row (c) is significant, whereas the interval of the 0.7\% gain of row (e) over row (d) includes zero.

\FloatBarrier
\section{Proofs}
\label{app:proofs}

\paragraph{Proof of Proposition~\ref{prop:poe}.}
By Bayes' rule and conditional independence,
\[
  p(c\mid h,a,b)\propto p(h,a,b\mid c)\,p(c)=p(h\mid c)\,p(a,b\mid c)\,p(c).
\]
Writing $p(h\mid c)=p(c\mid h)p(h)/p(c)$ and $p(a,b\mid c)=p(c\mid a,b)p(a,b)/p(c)$ and dropping factors that do not depend on $c$ gives $p(c\mid h,a,b)\propto p(c\mid h)\,p(c\mid a,b)/p(c)$. \qed

\paragraph{Proof of Proposition~\ref{prop:weight}.}
Write $\tilde q_k(c)=q_k(c)e^{\lambda T(c)}/Z(\lambda)$ over the candidates and the tail class, so $\log\tilde q_k(c^\star)=\log q_k(c^\star)+\lambda T(c^\star)-\log Z(\lambda)$.
Its first derivative in $\lambda$ is $T(c^\star)-\mathbb{E}_{\tilde q_k}T$ and its second derivative is $-\mathrm{Var}_{\tilde q_k}T\le0$, so the mean $L$ is concave, and setting $L'(\lambda^\star)=0$ gives the moment condition.
At $\lambda=0$, $\tilde q_k=q_k$, so the Newton step $-L'(0)/L''(0)$ is $\hat\lambda$.
If $c^\star\sim q_k$ given everything that $T$ depends on, then $\mathbb{E}[T(c^\star)\mid\cdot]=\mathbb{E}_{q_k}T$, so $L'(0)=0$, and concavity together with $\lambda\ge0$ gives $\lambda^\star=0$. \qed

\paragraph{Proof of Proposition~\ref{prop:opt}, part (a).}
A draft token at node $u$ is accepted exactly when the continuation starts with the path to $u$, which has probability $\tilde Q(u)$ under $\tilde q$.
Summing over nodes gives the expectation.
Each factor $\tilde q_k\le 1$, so $\tilde Q(\text{child})\le\tilde Q(\text{parent})$.
Ordering prefixes by $\tilde Q$ and breaking ties by depth, every prefix in the top $B$ has its parent in the top $B$, so the set is a valid tree.
No other valid tree with $B$ nodes can have a larger sum, since its sum is bounded by the sum of the $B$ largest values. \qed

\paragraph{Proof of Proposition~\ref{prop:opt}, part (b).}
Let $S^\star$ be the top-$B$ prefixes, which form a tree by part (a).
We show by induction over depth that every $u\in S^\star$ at depth $d$ is kept.
For $d=1$, $u$ has at most $B-1$ better prefixes, so at most $B-1$ better nodes at depth 1, and $u$ is among the top $W\ge B$ candidates.
For $d>1$, the parent of $u$ is in $S^\star$ and is kept by induction, so $u$ is a candidate at depth $d$, and the same counting argument shows that it is kept.
All of $S^\star$ is therefore among the kept nodes, and the $B$ best kept nodes are $S^\star$. \qed

\FloatBarrier
\section{Implementation Details}
\label{app:impl}
\begin{algorithm}[h]
\caption{DForkTree construction for one round}
\label{alg:build}
\begin{algorithmic}[1]
\small
\REQUIRE marginals $q_{1:K}$, candidates $\mathcal{C}_{1:K}$, $\mathcal{C}_0=\{r\}$, $\mathcal{C}_{-1}=\{x_{t-1}\}$, budget $B$, width $W\ge B$
\STATE Look up $T(a,b,c)$ for all $k\ge2$ and $(a,b,c)\in\mathcal{C}_{k-2}\times\mathcal{C}_{k-1}\times\mathcal{C}_k$
\STATE $L_1\leftarrow$ the $W$ best $c\in\mathcal{C}_1$ by $\log q_1(c)$
\FOR{$k=2,\dots,K$}
  \FOR{all $u\in L_{k-1}$ and $c\in\mathcal{C}_k$ \textbf{in parallel}}
    \STATE $\lambda\leftarrow\lambda_a$ if $u$ is on the spine, else $\lambda_p$
    \STATE $s(u,c)\leftarrow\log\tilde Q(u)+\log\tilde q_k(c\mid u)$ by \cref{eq:score}
  \ENDFOR
  \STATE $L_k\leftarrow$ the $W$ best pairs $(u,c)$ by $s(u,c)$
\ENDFOR
\STATE $\mathcal{T}\leftarrow\{r\}\,\cup$ the $B$ best nodes of $L_1\cup\dots\cup L_K$
\STATE \textbf{return} $\mathcal{T}$ with parent indices and ancestor mask
\end{algorithmic}
\end{algorithm}

\paragraph{Corpus.}
The corpus answers 100K prompts in seven categories, all drawn from public instruction datasets rather than from the evaluation benchmarks.
Applied math (20K) comes from the rephrased GSM8K training questions in MetaMathQA \citep{yu2024metamath} and from Orca-Math \citep{mitra2024orcamath}.
Competition math (15K) comes from the rephrased MATH training questions in MetaMathQA and the math subset of UltraInteract \citep{yuan2024ultrainteract}.
Competition code (20K) combines the coding subset of UltraInteract with OpenCodeReasoning \citep{ahmad2025opencodereasoning}, and function-level code (15K) comes from Evol-CodeAlpaca, an open reproduction of the code instructions of WizardCoder \citep{luo2024wizardcoder}.
Multi-turn chat (12K) keeps two or three user turns from UltraChat \citep{ding2023ultrachat}, instruction following (8K) uses AutoIF \citep{dong2025autoif}, and real user requests (10K) come from Chatbot Arena conversations \citep{chiang2024arena} and UltraFeedback \citep{cui2024ultrafeedback}, in English and Chinese only.
We keep only the prompts and discard every reference response.
Qwen3.5-4B answers each prompt once with thinking disabled, greedily for 60K prompts and with temperature 0.7, top-p 0.8 and top-k 20 for the other 40K.
In multi-turn chat it answers turn by turn after its own earlier answers, so every response token in the corpus is target text.
Only response tokens are counted.
Before generation, we drop prompts that share a 16-gram with any prompt of the seven benchmarks or that match a GSM8K or MATH test question exactly.
For LiveCodeBench we also drop the AtCoder and LeetCode problems of OpenCodeReasoning, its main sources, and the recently added Codeforces problems.
After generation, a check for shared 13-grams flags six near duplicates of test prompts.
Removing them leaves 99,994 responses and changes the throughput of DForkTree by $+0.2\%$ and that of DART by $-0.02\%$.

\paragraph{Count tables.}
We tokenize the corpus with the target tokenizer and count unigrams, bigrams and trigrams over each response.
With counts $n(\cdot)$, $N$ tokens and vocabulary $\mathcal{V}$, the smoothed estimates in \cref{eq:pmi} are
\[
  P(c) = \frac{n(c)+1}{N+|\mathcal{V}|}, \qquad
  P(c\mid b) = \frac{n(b,c)+\beta_2 P(c)}{n(b)+\beta_2}, \qquad
  P(c\mid a,b) = \frac{n(a,b,c)+\beta_3 P(c\mid b)}{n(a,b)+\beta_3},
\]
where $\beta_2=20$ and $\beta_3=60$ control how quickly higher-order counts are trusted.
Trigrams seen once are removed, and the context counts $n(a,b)$ are recomputed from the remaining trigrams so that the trigram estimate stays normalized.
Each table is stored on the GPU as a sorted array of packed integer keys with a parallel array of counts.
A lookup is a binary search over the keys, so the whole lattice of \cref{alg:build} is evaluated with one batched search per table.
For $K=23$ positions and $M=16$ candidates, the lattice holds about $(K-1)M^3\approx9\times10^4$ grandparent, parent and child combinations per round.
The tables hold 0.25M unigrams, 5.5M bigrams and 6.1M trigrams that occur at least twice.
Counting them over the 95.7M tokens of the 99,994 responses takes 27~s on one CPU thread, and their keys, counts and context counts take 163~MB of GPU memory.

\paragraph{Kernels.}
Each GDN layer updates its state with the gated delta rule $S_t=\alpha_t S_{t-1}(I-\beta_t k_t k_t^\top)+\beta_t v_t k_t^\top$ \citep{yang2025gated}.
The ancestor-masked chunk form uses $16\times16$ diagonal blocks, and the relative residual of the triangular solve stays near $2\times10^{-7}$ even for paths of depth 23.
Tiles are sized to fit the 99~KiB of shared memory per block on the RTX 5090, which matters at 48 and 64 nodes.
With $G_t$ the cumulative log-decay along the path to node $t$, $\tilde v_t$ the corrected values from the solve, $S_0$ the committed state and $n$ the last node of the accepted path $P$, unrolling the recurrence gives
\begin{equation}
  S_n=e^{G_n}S_0+\sum_{t\in P}e^{G_n-G_t}\,\tilde v_t k_t^\top.
  \label{eq:commit}
\end{equation}
This closed-form commit runs as one kernel over all GDN layers, heads and accepted nodes, and the convolution states are gathered from the inputs of the last three accepted nodes.
Attention planning for the tree mask is done on the device, so no host--device copy is needed inside a round.

\paragraph{Numerical checks.}
With a chain-shaped tree, our verifier reproduces the outputs of native DFlash verification.
On Qwen3.5-9B, the relative error of the fused tree GDN against an FP32 reference is 0.040, compared with 0.036 for the native recurrence.
The closed-form commit matches the replayed states with a maximum relative error of 0.010 for the recurrent states and 0.0036 for the convolution states.

\begin{figure}[h]
  \centering
  \includegraphics[width=0.5\textwidth]{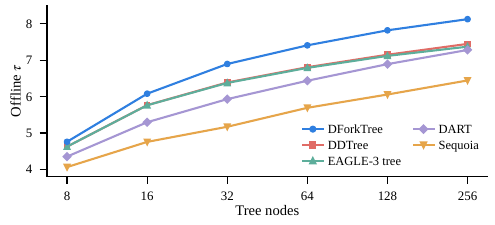}
  \vspace{-1.5mm}
  \caption{Offline $\tau$ of five construction strategies on Qwen3.5-4B with 23 draft positions, replayed along autoregressive outputs.}
  \label{fig:analysis}
\end{figure}

\paragraph{Offline replay.}
The replay in \cref{sec:budget} follows the autoregressive output of each A210 prompt.
At every round it builds the tree from cached drafter outputs (the top-64 candidates of each position for 24-token blocks), walks the tree along the reference output, and moves on by the accepted length plus one.
DDTree selects a candidate of rank 64 for at most 0.053\% of its nodes, so the cache does not truncate it.
The replayed $\tau$ differs from the online value by 0.08\% for DForkTree and 0.40\% for DDTree at 64 nodes.
\Cref{fig:analysis} compares five construction strategies in this replay.

\paragraph{Trees with 128 nodes.}
At 128 nodes the native state pool of SGLang needs about 12~GiB of scratch memory, so for both methods we raise the static memory fraction from 0.70 to 0.85 and keep every other setting.
Our kernels use 32-row tiles with $16\times16$ forward substitution and 64-bit state addressing at this size, and pass an FP64 reference check.

\FloatBarrier
\section{System Details}
\label{app:system}

\paragraph{Construction.}
With DDTree's marginal scores at 64 nodes and 23 positions, our layer-wise GPU search reaches the same acceptance as the official heap within 1\% ($\tau$ 7.72 vs.\ 7.67), as Proposition~\ref{prop:opt} predicts.
\Cref{tab:construct} times both on 100 cached rounds that start once the candidate scores are ready, with the heap followed by the GPU packing that all tree methods share.
The two cost about the same at 64 nodes.
The heap is faster for smaller trees and our search for larger ones, even though our search also evaluates the PMI.
Only the heap needs a round trip between GPU and host, which is part of the 1.6~ms build time that \cref{fig:system}a shows for DDTree.

\paragraph{Verification.}
\Cref{tab:verifier} runs each tree baseline with SGLang's native tree verification and with our verifier.
Our verifier speeds up all four tree methods by 13--17\% and shortens each round by 1.8--1.9~ms on Qwen3.5-4B, and it speeds them up by 10--12\% on Qwen3.5-9B.

\begin{table}[h]
\centering
\small
\begin{minipage}[t]{0.485\textwidth}
\centering
\setlength{\tabcolsep}{4pt}
\caption{Construction time per round in ms on cached rounds of Qwen3.5-4B with 23 positions, from candidate scores to the packed tree. Our search includes the PMI lookup.\label{tab:construct}}
\begin{tabular}{l cccc}
\toprule
Nodes & 16 & 32 & 64 & 128 \\
\midrule
DDTree heap and GPU packing & 0.21 & 0.25 & 0.41 & 0.57 \\
Layer-wise GPU search (ours) & 0.31 & 0.37 & 0.42 & 0.49 \\
\bottomrule
\end{tabular}
\end{minipage}\hfill
\begin{minipage}[t]{0.485\textwidth}
\centering
\setlength{\tabcolsep}{3pt}
\caption{Native SGLang tree verification versus our verifier on Qwen3.5-4B (848 prompts, one GPU, the same GPU packing in both columns). Throughput in tokens/s and time per round in ms.\label{tab:verifier}}
\begin{tabular}{l cc c cc}
\toprule
& \multicolumn{2}{c}{tok/s} & & \multicolumn{2}{c}{ms / round} \\
\cmidrule(lr){2-3}\cmidrule(lr){5-6}
Method & native & ours & gain & native & ours \\
\midrule
Sequoia tree & 429 & 501 & +16.8\% & 12.83 & 10.92 \\
DART tree & 386 & 437 & +13.2\% & 15.38 & 13.59 \\
EAGLE-3 tree & 480 & 557 & +16.0\% & 13.08 & 11.19 \\
DDTree & 461 & 532 & +15.4\% & 13.56 & 11.76 \\
\bottomrule
\end{tabular}
\end{minipage}
\end{table}

\paragraph{Microbenchmark.}
\Cref{fig:system}c isolates the short convolution and the 24 GDN layers.
Native verification takes 1.3, 2.5, 5.4, 12.2 and 26.7~ms at 16, 32, 64, 128 and 256 nodes and writes 0.77 to 12.3~GiB of state snapshots per round.
Our kernels take 0.26, 0.42, 1.8, 2.8 and 7.9~ms.
At 256 nodes our kernels tile the solve in blocks of 64 columns to stay within shared memory.

\FloatBarrier
\section{Baselines}
\label{app:baselines}
All tree baselines take the top-16 candidates of the same DFlash drafter, use blocks of 16 draft tokens and 32 nodes, and run on the verifier of \cref{sec:verify}.
They also share DForkTree's packing of the tree on the GPU, which computes depths, ancestor masks and child and sibling links without reading device values on the host.
Sequoia gathers its static tree and the EAGLE-3 tree expands and ranks its candidates on the GPU, both inside CUDA graphs.
DDTree and DART run their official searches on the host and pass only tokens and parents to the GPU packer.
The MTP baseline uses Qwen3.5's built-in head with 4 steps and top-4.
On 400 cached rounds and 14 prompts, these implementations select the same nodes and produce the same outputs as direct ports of the original code, so the baselines differ only in how the tree is built.

\paragraph{Sequoia.}
We estimate the probability that the candidate of rank $j$ is accepted given that its parent is accepted, from offline replays of drafter outputs, pooled over depths.
Sequoia's dynamic program then finds the static tree with the largest expected acceptance under this vector.
The vector is calibrated again for Qwen3.5-9B.

\paragraph{DART.}
We use the official C++ tree search with its default settings, namely 25 candidates per position, beam width 20, level weights $(l+1)^{-0.7}$, logit weights $0.9^{l}$, an $n$-gram weight of 0.5 and no $n$-gram score at the first position.
The $n$-gram trie is built from the same corpus as DForkTree.
The official encoder can produce an invalid tree when two scores tie exactly, and we break such ties by index, which changes only the affected rounds.

\paragraph{EAGLE-3 tree.}
We follow the dynamic tree of EAGLE-2, which EAGLE-3 also uses.
At each depth the 10 nodes with the largest cumulative log-probability are expanded with their top-10 children, and the final tree keeps the best nodes by cumulative score up to the budget, closed under ancestors.
We use all 15 draft positions, since a block diffusion drafter produces them in one pass.

\paragraph{DDTree.}
We use the official best-first heap on the drafter's marginals.

The budgets and block sizes of all tree methods were selected from $\{16,32,64\}$ nodes and blocks of $\{16,24\}$ tokens on a 21-prompt development set before the test runs, which gave 32 nodes for all methods, blocks of 16 for the baselines and of 24 for DForkTree.
For the offline replay in \cref{sec:budget}, Sequoia's acceptance vector is calibrated for 24-token blocks on the first three A210 prompts of each benchmark, which are part of the analysis set.

\FloatBarrier
\section{Corpus Size, Domain and Source}
\label{app:corpus}
DForkTree needs a corpus of target generations, so we measure how its benefit depends on that corpus.
The offline numbers below replay the 155 development trajectories with blocks of 16 tokens and 64 nodes, and the online numbers use A210 with 64 nodes and 23 positions.

\begin{figure}[h]
  \centering
  \begin{minipage}{0.85\textwidth}
    \centering
    \includegraphics[width=\linewidth]{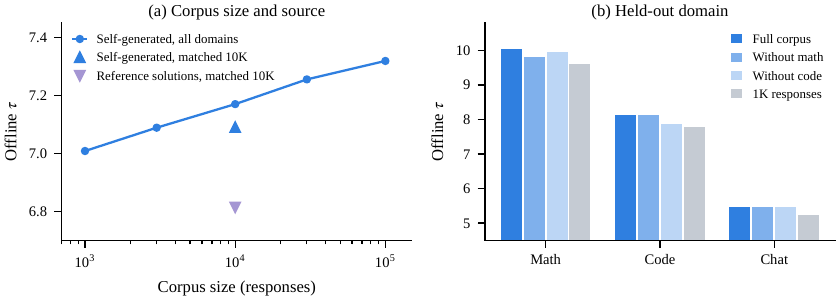}
    \caption{(a) Offline $\tau$ as the corpus shrinks, and with 10K matched math and code prompts answered by the target or by published reference solutions. (b) Offline $\tau$ pooled by domain when the corpus excludes all math or all code responses.}
    \label{fig:corpus}
  \end{minipage}
\end{figure}

\paragraph{Size.}
We subsample the 100K responses with a fixed seed into nested subsets.
In \cref{fig:corpus}a, $\tau$ grows roughly linearly in the logarithm of the corpus size, from 7.01 with 1K responses to 7.32 with 100K, and has not saturated.
Online on A210, 1K, 10K and 100K responses give 532, 543 and 555 tokens/s, against 514 tokens/s for DDTree in the same configuration.

\paragraph{Source.}
For 10K math and code prompts we take published, correctness-filtered reference solutions to the same prompts from UltraInteract \citep{yuan2024ultrainteract} and count them with the same settings.
Self-generated text gives a higher $\tau$ on every benchmark (7.09 vs.\ 6.81 overall), although it also contains 2.7 times as many tokens (9.1M vs.\ 3.4M).

\paragraph{Domain.}
\Cref{fig:corpus}b removes all math or all code responses, about 35K each.
On the held-out domain $\tau$ drops by 2--3\%, which still beats a 1K corpus that includes the domain, and the other domains change by less than 1\%.

\FloatBarrier

\section{Sensitivity and Variants}
\label{app:sens}
\Cref{tab:sens} replays DForkTree on the 155 development trajectories with blocks of 16 tokens and 64 nodes.
Because these trajectories were used to select the hyperparameters, the table describes sensitivity and does not estimate held-out gains.
The selected point is the best of the grid, and $\beta_3$ hardly matters.
The best single weight, $\lambda_a=\lambda_p=0.3$, is 1.4\% lower, and reversing the order of the two weights is worse than either single weight.
A 4-gram level on top of the trigram counts and a prior at the first position both lower $\tau$ slightly.

\begin{table}[h]
\centering
\small
\fittab{Offline $\tau$ on 155 development trajectories (blocks of 16, 64 nodes).\label{tab:sens}}{%
\begin{tabular}{l ccc c}
\toprule
Setting & $\lambda_a$ & $\lambda_p$ & $\beta_3$ & $\tau$ \\
\midrule
Selected & 0.3 & 0.6 & 60 & 7.318 \\
Smoothing & 0.3 & 0.6 & 20 / 180 & 7.313 / 7.316 \\
Single weight & 0.3 & 0.3 & 60 & 7.216 \\
Single weight & 0.6 & 0.6 & 60 & 7.169 \\
Reversed order & 0.6 & 0.3 & 60 & 7.094 \\
No prior on the spine & 0 & 0.6 & 60 & 7.167 \\
Weaker spine weight & 0.15 & 0.6 & 60 & 7.309 \\
Stronger branch weight & 0.3 & 0.9 / 1.2 & 60 & 7.274 / 7.212 \\
4-gram level ($\beta_4=60$) & 0.3 & 0.6 & 60 & 7.309 \\
Prior at the first position & 0.3 & 0.6 & 60 & 7.291 \\
\bottomrule
\end{tabular}}
\end{table}

\begin{figure}[h]
  \centering
  \begin{minipage}[t]{0.485\textwidth}
    \centering
    \includegraphics[width=\linewidth]{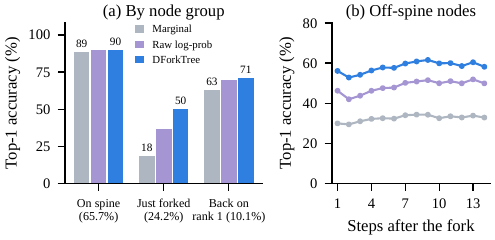}
    \caption{(a) Top-1 accuracy of the drafter's marginal, of the marginal corrected by the raw $n$-gram log-probability as in DART, and of DForkTree's score, by node group. Parentheses give each group's share of the 1.03M events. (b) Top-1 accuracy of off-spine nodes by the number of steps since the path left the spine.}
    \label{fig:cohort}
  \end{minipage}\hfill
  \begin{minipage}[t]{0.485\textwidth}
    \centering
    \includegraphics[width=\linewidth]{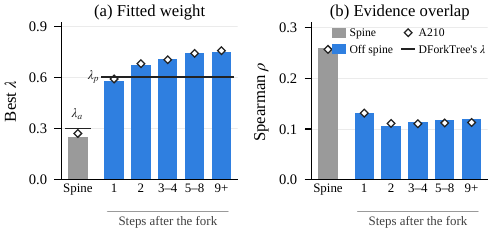}
    \caption{Fitted weight and evidence overlap by the number of steps since the parent path left the spine. Bars show development replays and diamonds show A210. (a) Best $\lambda$ per group, with DForkTree's two weights as black lines. (b) Spearman correlation between the drafter's log-probability and the PMI over the candidates of a position.}
    \label{fig:lambdamain}
  \end{minipage}
\end{figure}

\begin{figure}[h]
  \centering
  \begin{minipage}{0.8\textwidth}
    \centering
    \includegraphics[width=\linewidth]{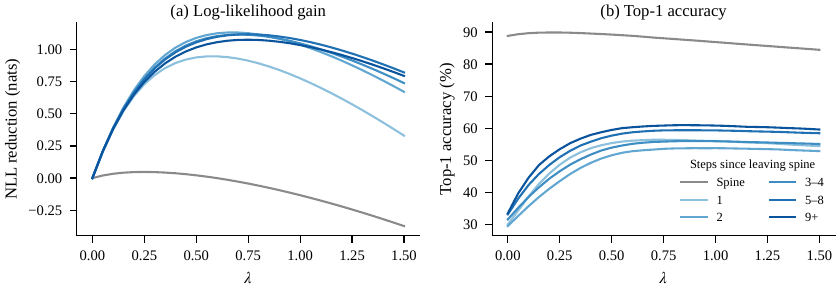}
    \caption{Development replays grouped by the number of steps since the parent path left the spine. (a) Reduction in the negative log-likelihood of the true next token as a function of $\lambda$. (b) Top-1 accuracy of the re-scored candidates as a function of $\lambda$.}
    \label{fig:lambda}
  \end{minipage}
\end{figure}

\paragraph{Per-group fit.}
\Cref{fig:cohort} compares the three scores by node group and follows off-spine nodes along the branch.
\Cref{fig:lambdamain} summarizes the weights fitted per group, and \cref{fig:lambda} shows the curves behind it.
Each group contains the reachable positions $k\ge2$ whose parent path has the given relation to the spine, 1.03M events in total, and targets outside the top-16 candidates count with probability $\rho_k/Z$.
On the spine the prior lowers the negative log-likelihood by only 0.05 nats at its best weight and slightly raises top-1 accuracy from 88.8\% to 89.9\%.
After the fork it lowers the negative log-likelihood by about 1.1 nats and raises top-1 accuracy from 31.4\% to 56.8\%.
Among off-spine nodes whose parent is again a top-1 token, the best weight is 0.45, between the two groups.
The difference between the off-spine and spine weights is 0.43 (95\% CI 0.42--0.45) on the development replays and 0.42 (0.41--0.43) on A210.

\paragraph{Weight and overlap.}

\begin{figure}[h]
  \centering
  \begin{minipage}{0.6\textwidth}
    \centering
    \includegraphics[width=0.82\linewidth]{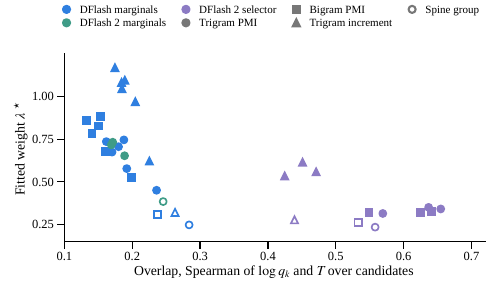}
    \caption{Overlap of the evidence with the base score against its fitted weight, for the 37 groups of \cref{fig:weighthat}. The overlap is the mean Spearman correlation between the base log-probability and the evidence over the candidates of a position.}
    \label{fig:weight}
  \end{minipage}
\end{figure}

We test Proposition~\ref{prop:weight} on 37 groups of nodes, on the development replays and on A210.
For DFlash they are the seven groups above, each with the trigram PMI, the bigram PMI and the trigram increment as evidence.
For DFlash 2 they are the spine and the first three tokens after a fork, with the trigram PMI on its marginals and with the bigram PMI, the trigram PMI and the increment on its selector, which is given the tail mass as in the full-tree comparison below.
Across groups, $\hat\lambda$ and $\lambda^\star$ have Spearman correlations of 0.89 and 0.87 on the two sets.
The fitted weight falls with the mean correlation between $\log q_k$ and $T$ ($-0.78$ on both sets) and with the share of $T$ that $\log q_k$ explains ($-0.77$ and $-0.69$).
On top of the selector, the bigram PMI receives 0.26--0.33 at every distance, whereas the increment rises from 0.28 on the spine to 0.53--0.62 after a fork.
The groups share events, so these correlations describe a pattern rather than independent replicates.
\Cref{fig:weighthat,fig:weight} also show what the overlap does not explain.
Along a branch of DFlash, $\lambda^\star$ for the trigram PMI rises from 0.58 at the first token after a fork to 0.75 five or more tokens later.
The overlap stays near 0.18 over this range and $\hat\lambda$ does not rise, so deeper tokens need more weight than the first-order estimate suggests.
The trigram increment on DFlash's marginals also receives weights near 1 after a fork, because its values span a smaller range than those of the full PMI.

\paragraph{Mechanism checks.}
Further replays test the explanation of \cref{sec:analysis}.
Right after the first fork, the drafter's top-1 accuracy rises from 18\% to 48\% with the marginal probability of the fork token itself.
This is what we would expect if each marginal averages over the drafter's own predecessors, although DForkTree's gain does not shrink as that probability grows.
The trigram's advantage over the bigram grows along the branch, from 0.06 nats and 2.0 points of top-1 accuracy at the first token after a fork to 0.12 nats and 3.4 points five or more tokens later.
In the trees of \cref{sec:sources}, DForkTree lowers the expected calibration error of the path scores from 0.092 to 0.074 on spine nodes and from 0.0038 to 0.0033 on off-spine nodes.

\paragraph{Further variants.}
Several further training-free variants do not help.
Replacing $P(c)$ in \cref{eq:pmi} by the trigram mixture over the drafter's candidates at the previous position, weighted by their marginals, changes offline $\tau$ by $-0.2\%$ and $+0.2\%$ at 32 and 64 nodes.
The weights fitted per group still differ between spine and off-spine nodes, at 0.40 against 0.80.
Splitting the branch weight into one for the first token after a fork and one for deeper tokens raises offline $\tau$ on A210 by at most 0.4\% at 32 and 64 nodes, with or without a four-gram level whose weights are fitted for each group.
None of these variants decodes faster than DForkTree's own scores in the same implementation on the development prompts.
Adding counts from the current prompt and output to the corpus raises offline $\tau$ by only 0.3\%.
Choosing each round's budget is studied with calibrated scores below.

\paragraph{Branch tokens without target features.}
Re-anchoring gives the drafter both the branch tokens and the target's features of them.
To separate the two, we fill the branch tokens into the drafter's block in place of the masks and keep the context features of the committed prefix.
At the first three tokens after a fork, this raises the drafter's top-1 accuracy from 30--31\% to 37--39\%, against 54--56\% for DForkTree and 86--90\% for the full re-run.
Replacing DForkTree's scores with these predictions up to one or three tokens after a fork in the trees of \cref{sec:sources} even lowers offline $\tau$, to 6.79 and 6.72 against 6.96 at 32 nodes.
DFlash is trained with masks after the first position of a block, so these inputs lie outside its training distribution, and a drafter trained to read branch tokens may behave differently.
For the released drafter, the knowledge that re-anchoring reveals is thus reached through the target's features of the branch.
An extra drafter pass over the branch tokens does worse than DForkTree's counts.

\paragraph{A learned selector.}
DFlash 2 \citep{dflash2blog2026} adds to the DFlash drafter a trained selector that scores each candidate given its predecessor with a low-rank bilinear term.
We use community weights for Qwen3.5-4B, trained against a quantized target, and define the spine and the forks on DFlash 2's own marginals with blocks of 16.
At the first three tokens after a fork on A210, the PMI correction of DForkTree closes 44--45\% of the gap between DFlash 2's marginals and DFlash 2 re-run at the fork.
The trained selector closes 46--47\%.
Adding the PMI to the selector, with a weight of 0.20--0.30 fitted on the development replays instead of 0.55--0.60 on the marginals, closes 49--50\%.
Splitting the PMI into its bigram part $\log P(c\mid b)/P(c)$ and its trigram increment $\log P(c\mid a,b)/P(c\mid b)$ shows where this gain comes from.
On top of the selector, the bigram part adds only 0.5--1.7 points of closure and receives a weight of 0.15--0.20.
The increment adds 3.5--4.2 points with a weight of 0.45--0.55 and does slightly better than the full PMI.
The selector and the bigram both condition on the parent, so the counts add mainly the token before the parent, which a first-order selector does not see.
The fitted weight again follows how much the evidence overlaps with what the scorer already knows, as on the spine in \cref{sec:fork}.
The gain is largest for trigram contexts of medium frequency, where counts are reliable but not so frequent that the drafter already reflects them.
At the node level, the trigram increment alone is therefore the evidence to add to a learned first-order module.

\paragraph{Full trees with DFlash 2.}
In full trees the order reverses.
We use blocks of 16, the same exact search for every score and weights selected on the development replays.
On A210, DForkTree's scores on DFlash 2's marginals then raise offline $\tau$ by 10.9\% at 32 nodes and 11.6\% at 64 nodes, against 8.8\% and 9.6\% for the trained selector.
DForkTree thus accepts 1.9\% and 1.8\% more tokens per round (95\% CIs 1.4--2.4\% and 1.3--2.3\%), and adding the trigram increment to the selector gains only 0.7\%.
Two controlled variants locate the gap. Giving the selector the tail mass that DForkTree keeps raises its $\tau$ by 0.7\%, a third of the gap, and removes its overconfidence on branch nodes.
Without the tail mass, the selector's path scores at 32 nodes predict an acceptance rate of 8.7\% for branch nodes against an actual 5.4\%, while DForkTree predicts 3.9\% against 5.6\%.
Restoring the marginals for the children of spine nodes instead lowers $\tau$ by 5.7\%.
Most of the selector's gain therefore comes from re-ranking the candidates where a branch leaves the spine, and DForkTree's prior does this at least as well.
Replacing DForkTree's scores on branches by the selector with tail mass and the increment changes $\tau$ by less than 0.1\%, so in full trees the selector adds nothing to DForkTree's counts.

\paragraph{Candidates after a fork.}
All scores above re-rank the drafter's top 16 candidates at a position, and these candidates are computed for the spine too.
On A210 the true next token is among them for 98.3\% of spine nodes but for only 80.1\%, 74.0\% and 73.9\% of the nodes one, two and three tokens after a fork.
Even the top 64 miss 9.5--12.3\% of the latter.
Any re-ranking of the 16 candidates is capped at these rates, which accounts for 4\%, 18\% and 21\% of the gap to the re-anchored drafter at the three steps.
The rest lies in the ranking, where DForkTree reaches 45--47\% of the room between the marginals and the cap and the learned head below reaches 51--52\%.
The eight most frequent trigram continuations of the grandparent and parent that are not among the candidates contain 37--41\% of the missing tokens after a fork and raise the coverage to 84--87\%.
This is 3.3--5.8 points above the same number of extra candidates from the drafter's own ranking, whereas on the spine they add nothing.

\paragraph{Candidates by position.}
\Cref{fig:motivation}c follows the target's true continuation from every anchor of A210 with 23 draft positions.
It counts a position only while the true tokens at all earlier positions are among their top 16 candidates, which is the condition under which DFlash 2 reports coverage \citep{dflash2blog2026}.
A position is on the spine if all earlier true tokens are top-1 tokens, and off the spine otherwise.
With $S_k$ and $F_k$ the coverage on and off the spine at position $k$, $p_k$ the share of positions still on the spine and $p_1=1$, the pooled coverage $R_k=p_kS_k+(1-p_k)F_k$ falls from position 1 to $K$ by
\[
  R_1-R_K=(S_1-S_K)+(1-p_K)(S_K-F_K),
\]
where the first term is the decline on the spine and the second the growing share of paths that have left it.
\Cref{tab:pa21} shows that the second term accounts for 90--95\% of the decline for both drafters, while coverage on the spine stays above 96\% at the end of the block.
On the GSM8K prompts of A210 the share is 89\% for DFlash and 78\% for DFlash 2 at $K=7$.
Coverage after a fork also falls with the rank of the fork token.
For DFlash it is 86.9\% after a fork to the second candidate, 81.4\% after the third or fourth and 74.2\% after a lower one, and for DFlash 2 the same numbers are 83.6\%, 79.2\% and 72.9\%.
Top-1 accuracy on the spine does fall in the deepest positions, from about 90\% to 75\% between positions 15 and 23 for DFlash, so deep on the spine the true token stays among the candidates while the first choice weakens.
Because the condition drops paths whose earlier true tokens left the top 16, off-spine coverage here is 4.7--8.3 points higher than for the tree nodes of \cref{sec:analysis}.

\begin{table}[h]
\centering
\small
\fittab{Decomposition of the decline of top-16 coverage between position 1 and position $K$ on A210, in percentage points, with 95\% CIs of the share from a prompt-clustered bootstrap. The share can exceed 100\% when the two terms have opposite signs.\label{tab:pa21}}{%
\begin{tabular}{l c ccc c c}
\toprule
Drafter & $K$ & Decline & On the spine & Leaving the spine & Share of the second term & Spine coverage at $K$ (\%) \\
\midrule
DFlash & 7 & 7.3 & 0.4 & 7.0 & 95\% (92--98) & 98.9 \\
DFlash & 23 & 9.5 & 0.8 & 8.8 & 92\% (88--96) & 98.5 \\
DFlash 2 & 7 & 10.3 & 1.0 & 9.3 & 90\% (86--95) & 96.7 \\
DFlash 2 & 15 & 10.8 & 0.6 & 10.2 & 95\% (88--103) & 97.1 \\
\bottomrule
\end{tabular}}
\end{table}

\paragraph{Tree shape.}
\Cref{tab:shape} gives the shape of the replayed trees of \cref{sec:sources}.
At 64 nodes, DDTree and DForkTree keep a similar number of spine, off-spine and branching nodes, while DForkTree's branches reach about two levels deeper.

\begin{table}[h]
\centering
\caption{Tree shape per round (replay, mean over rounds). Branching nodes have two or more children.}
\label{tab:shape}
\small
\setlength{\tabcolsep}{4pt}
\begin{tabular}{l ccc}
\toprule
& \multicolumn{2}{c}{DDTree} & DForkTree \\
\cmidrule(lr){2-3}\cmidrule(lr){4-4}
Nodes / block & 32 / 16 & 64 / 24 & 64 / 24 \\
\midrule
Spine nodes & 7.3 & 8.3 & 7.9 \\
Off-spine nodes & 23.7 & 54.7 & 55.1 \\
Branching nodes & 5.3 & 10.2 & 10.6 \\
Maximum depth & 7.3 & 8.3 & 10.3 \\
Rounds leaving the spine (\%) & 58.6 & 71.7 & 74.2 \\
\bottomrule
\end{tabular}
\end{table}

\paragraph{Calibration by node type.}
We replay DDTree, the EAGLE-3 tree and DForkTree with 23 draft positions on A210 and compare each node's path score with whether the target's true continuation passes through it.
\Cref{tab:cal0} shows that all three underestimate acceptance on the spine and at the fork node and overestimate it after a fork.
Pooled over each round, the three trees predict 0.88--0.91 of the draft tokens they accept at 32 nodes and 0.89--0.91 at 64 nodes.
The spine carries most of the probability and hides the opposite error after the fork.
DForkTree's scores leave a slightly smaller error on the spine and remove most of the overestimate after the fork.

\begin{table}[h]
\centering
\small
\fittab{Mean path score and acceptance rate by node type in trees with 32 nodes on A210, with their ratio. The fork node is where a path first leaves the spine.\label{tab:cal0}}{%
\begin{tabular}{l ccc ccc ccc}
\toprule
& \multicolumn{3}{c}{Spine} & \multicolumn{3}{c}{Fork node} & \multicolumn{3}{c}{After the fork} \\
\cmidrule(lr){2-4}\cmidrule(lr){5-7}\cmidrule(lr){8-10}
Method & score & accepted & ratio & score & accepted & ratio & score & accepted & ratio \\
\midrule
DDTree & 0.474 & 0.570 & 1.20 & 0.051 & 0.058 & 1.13 & 0.050 & 0.035 & 0.71 \\
EAGLE-3 tree & 0.458 & 0.551 & 1.20 & 0.052 & 0.059 & 1.13 & 0.049 & 0.035 & 0.72 \\
DForkTree & 0.528 & 0.604 & 1.15 & 0.067 & 0.073 & 1.09 & 0.064 & 0.060 & 0.94 \\
\bottomrule
\end{tabular}}
\end{table}

\paragraph{Calibrated scores.}
The opposite errors of \cref{tab:cal0} suggest giving each kind of node its own temperature.
We raise the drafter's probabilities of the candidates and of the tail to a power $\beta_g$ before the normalization of \cref{eq:score}.
The eight groups $g$ are given by whether the parent path is on the spine and by four depth ranges.
We fit the $\beta_g$ on the development replays in three ways, by the edge likelihood, by the path calibration error from shallow to deep depths, or by offline $\tau$ directly.
\Cref{tab:cal1} shows that one global temperature changes almost nothing, and that per-group temperatures remove most of the post-fork overestimate for both methods but raise offline $\tau$ by at most 1.1\%.
DDTree gains the most, but DForkTree still leads calibrated DDTree by 6.8\% at 32 nodes.
DForkTree's gain thus comes mainly from re-ranking the candidates of a branch rather than from rescaling its scores.
Fitting to the path calibration or to $\tau$ beats fitting to the edge likelihood by 0.030 and 0.011 in $\tau$ at 32 nodes (95\% CIs 0.015--0.046 and 0.001--0.022), a small effect in the direction of \cref{sec:ceiling}.
We also choose each round's budget among 16, 24, 32, 48 and 64 nodes from the path scores, using measured times per round.
Compared with a fixed 32 nodes, this changes estimated throughput by $-0.3\%$ with raw scores, by $-0.3\%$ with one global rescaling and by $+0.3\%$ with the per-group temperatures, against $+4.9\%$ for an oracle that knows each round's acceptance.
On 21 development prompts, the temperatures fitted to $\tau$ did not raise online throughput ($-2.8\%$, 95\% CI $-5.1$ to $-0.7\%$), so DForkTree keeps its uncalibrated scores.

\begin{table}[h]
\centering
\small
\fittab{Offline $\tau$ on A210 with temperatures per node group fitted on the development replays, and the ratio of acceptance rate to path score for nodes after a fork at 32 nodes. Changes are relative to the same method without calibration.\label{tab:cal1}}{%
\begin{tabular}{l l cc cc c}
\toprule
Method & Temperatures fitted by & $\tau$, 32 nodes & change & $\tau$, 64 nodes & change & Post-fork ratio \\
\midrule
DDTree & none & 6.376 & -- & 6.789 & -- & 0.71 \\
DDTree & one global temperature & 6.391 & $+0.23\%$ & 6.797 & $+0.11\%$ & 0.70 \\
DDTree & edge likelihood & 6.447 & $+1.11\%$ & 6.840 & $+0.76\%$ & 0.95 \\
DForkTree & none & 6.887 & -- & 7.399 & -- & 0.94 \\
DForkTree & edge likelihood & 6.911 & $+0.35\%$ & 7.431 & $+0.44\%$ & 1.00 \\
DForkTree & offline $\tau$ & 6.922 & $+0.51\%$ & 7.447 & $+0.65\%$ & 1.02 \\
DForkTree & path calibration & 6.941 & $+0.79\%$ & 7.454 & $+0.75\%$ & 1.03 \\
\bottomrule
\end{tabular}}
\end{table}

\paragraph{A learned re-anchoring head.}
We also distill the re-anchored drafter into a small head on top of the frozen DFlash drafter.
The head reads the drafter's hidden state at the position, 256-dimensional embeddings of the parent, the grandparent and the spine token that the drafter assumed there, and the distance from the fork.
From these it adds a bilinear correction to the scores of the 16 candidates and keeps the tail mass.
We train it on 2.5M nodes from 3{,}000 responses held out from the corpus, with the re-anchored drafter as the teacher, and recount the $n$-grams without these responses for every method.
In this setting DForkTree closes 45\%, 37\% and 36\% of the gap at the first three tokens after a fork on A210.
The head alone closes 38\%, 34\% and 34\%, and with the PMI as an additional input it closes 49\%, 43\% and 41\%, keeps the spine unchanged and halves the calibration error on branches.
Without the drafter's hidden state, it loses only 1--2 points.
By the oracle of \cref{sec:ceiling}, five more points of closure would lengthen acceptance by only about 1\%, less than a head of this size likely costs per round.
The head sees far less data than the counts, so a larger model or corpus may close more.
Within this budget, however, the drafter's own state carries little of the missing information, which agrees with the test above in which branch tokens alone did not help.

\paragraph{A second drafter pass.}
Re-anchoring needs the target's features of the branch, but a drafter trained to read branch tokens might recover part of the gain in a second pass.
We fine-tune a LoRA adapter (rank 64, 21.4M parameters) on all linear layers of the frozen DFlash drafter for one epoch on 3{,}000 responses held out from the corpus.
Given the context features up to the anchor and the true tokens of a path, the adapted drafter predicts the next eight tokens.
In each round of the replay, the adapter re-drafts from the $K$ best branch nodes up to depth four in DForkTree's tree.
The candidates and scores below these nodes then come from this second pass, with the PMI added as before.
On A210, offline $\tau$ rises by 1.0--2.0\% at 32 nodes and by 1.4--2.9\% at 64 nodes for $K$ from 1 to 8.
A CUDA-graph benchmark puts the second pass at 2.5--6.2~ms on Qwen3.5-4B and 3.6--7.7~ms on Qwen3.5-9B, depending on $K$ and on contexts of 256 to 2048 tokens.
Against rounds of 12 and 19~ms, it would need 21--52\% and 18--40\% longer acceptance to pay off.
The replay does not separate the effect of training from that of the second pass itself, and we leave an online test to future work.

\FloatBarrier
\section{Limitations}
\label{app:limits}
All our measurements use batch size one on RTX 5090 GPUs.
Larger batches leave less idle computation for verification, so the gains of all tree methods, including DForkTree, may change under batched serving, which we have not measured.
DForkTree also needs a corpus of target generations, which the target produces once, offline.
With only 1K responses, DForkTree already reaches 532 tokens/s on A210 against 514 for DDTree (Appendix~\ref{app:corpus}), and Qwen3.5-9B reuses the corpus of Qwen3.5-4B.
Our analysis applies to any drafter whose positions are predicted independently, but we test it only on block diffusion drafters.
Finally, both target models come from the Qwen3.5 family, and we test DFlash 2 only in offline replays with community weights trained against a quantized target.

\end{document}